\documentclass[10pt,journal,compsoc]{IEEEtran}

\usepackage{times}
\usepackage[utf8]{inputenc} 
\usepackage[T1]{fontenc}    
\usepackage[pagebackref=true,breaklinks,colorlinks,citecolor=blue,linkcolor=blue,urlcolor=blue,hypertexnames=false]{hyperref}
\usepackage{url}            
\usepackage{booktabs}       
\usepackage{amsfonts}       
\usepackage{nicefrac}       
\usepackage{microtype}      
\usepackage{xcolor}         
\usepackage{makecell}
\usepackage{adjustbox}
\usepackage{pifont}
\usepackage{bbding}
\usepackage{subcaption}
\usepackage{tabulary,multirow,xspace}
\usepackage{fixmath,mathtools,nicefrac}
\usepackage{colortbl}
\usepackage{xcolor}
\definecolor{mygray}{gray}{0.95}
\usepackage[misc]{ifsym} 
\usepackage{float}
\usepackage{array}
\usepackage{makecell}
\usepackage{ulem}
\usepackage{ragged2e}
\usepackage{bm}
\usepackage{algorithm, algorithmic}
\usepackage{arydshln}
\usepackage{enumitem}

\usepackage{graphicx} 
\usepackage{multirow}
\usepackage{caption}
\usepackage{balance}
\usepackage{makecell}

\newcommand{\model}{X-Fusion~Net}
\newcommand{\modelname}{X-Fusion}

\ifCLASSOPTIONcompsoc
  \usepackage[nocompress]{cite}
\else
  \usepackage{cite}
\fi
\ifCLASSINFOpdf
\else
\fi
\usepackage{amsmath}
\begin{document}

\title{Human-in-Context: Unified Cross-Domain 3D Human Motion Modeling via In-Context Learning}

\author{Mengyuan Liu,
        Xinshun Wang$^{\dagger}$,
        Zhongbin Fang$^{\dagger}$,
        Deheng Ye,
        Xia Li,
        Tao Tang,
        Songtao Wu,
        Xiangtai Li,
        Ming-Hsuan Yang
        
\IEEEcompsocitemizethanks{
\IEEEcompsocthanksitem Mengyuan Liu, Xinshun Wang, and Tao Tang are with the National Key Laboratory of General Artificial Intelligence, Peking University, Shenzhen Graduate School.
\IEEEcompsocthanksitem Zhongbin Fang and Deheng Ye are with Tencent, China.
\IEEEcompsocthanksitem Xia Li is with the Department of Information Technology and Electrical Engineering, ETH Zurich.
\IEEEcompsocthanksitem Xiangtai Li is with S-Lab, Nanyang Technological University, Singapore.
\IEEEcompsocthanksitem Songtao Wu is with Sony R\&D Center, China.
\IEEEcompsocthanksitem Ming-Hsuan Yang is with the University of California at Merced, US.
\IEEEcompsocthanksitem Xinshun Wang and Zhongbin Fang are co-corresponding authors.
}
}

%
%


\IEEEtitleabstractindextext{
\begin{abstract}
\justifying
This paper aims to model 3D human motion across domains, where a single model is expected to handle multiple modalities, tasks, and datasets. 
Existing cross-domain models often rely on domain-specific components and multi-stage training, which limits their practicality and scalability. 
To overcome these challenges, we propose a new setting to train a unified cross-domain model through a single process, eliminating the need for domain-specific components and multi-stage training. 
We first introduce Pose-in-Context (PiC), which leverages in-context learning to create a pose-centric cross-domain model. 
While PiC generalizes across multiple pose-based tasks and datasets, it encounters difficulties with modality diversity, prompting strategy, and contextual dependency handling. 
We thus propose Human-in-Context (HiC), an extension of PiC that broadens generalization across modalities, tasks, and datasets. HiC combines pose and mesh representations within a unified framework, expands task coverage, and incorporates larger-scale datasets. 
Additionally, HiC introduces a max-min similarity prompt sampling strategy to enhance generalization across diverse domains and a network architecture with dual-branch context injection for improved handling of contextual dependencies. 
Extensive experimental results show that HiC performs better than PiC in terms of generalization, data scale, and performance across a wide range of domains. 
These results demonstrate the potential of HiC for building a unified cross-domain 3D human motion model with improved flexibility and scalability. 
The source codes and models are available at \textcolor{blue}{https://github.com/BradleyWang0416/Human-in-Context}.
\end{abstract}

}

\maketitle

\IEEEdisplaynontitleabstractindextext

%
\IEEEpeerreviewmaketitle

\IEEEraisesectionheading{\section{Introduction}\label{sec:introduction}}

\IEEEPARstart{A}s a core topic in computer vision, cross-domain 3D human motion modeling aims to develop a model capable of handling multiple domains, including different tasks, modalities, and datasets. 
To achieve this, poses~\cite{liu2017enhanced,chen20173d} and meshes~\cite{choi2020pose2mesh} are two widely adopted human motion representations, as they provide greater efficiency, compactness, and richer information compared to RGB-based representations~\cite{liu2018pem,liu2025milnet}. 
Leveraging pose and mesh representations, cross-domain 3D human motion modeling has found diverse applications, such as motion prediction for autonomous driving~\cite{wang2024dynamic,li2021symbiotic}, pose estimation for human-robot collaboration~\cite{liu2025tcpformer,li2024hot,liu2019feature}, and mesh recovery for virtual reality~\cite{tang2024arts}.

Cross-domain models have been extensively explored in computer vision, robotics, and natural language processing, leveraging diverse backbones~\cite{vaswani2017attention,dosovitskiy2020image,zhu2024vision} and training paradigms~\cite{devlin2018bert,brown2020language}. 
However, applying cross-domain modeling to 3D human motion presents even greater challenges due to the multi-dimensional and spatiotemporal complexity of 3D motion data~\cite{yan2018spatial}.  
As a result, existing cross-domain models~\cite{zhu2023motionbert,foo2023UPS,zhang2025lmm,ci2023unihcp,wu2024macdiff} suffer from two major limitations. 
First, their applicability is often restricted to a narrow scope, such as performing the same task across a few datasets~\cite{mao2020history,wang2024gcnext} or handling similar tasks within single-modal data~\cite{foo2023UPS,cui2021towards}. 
Second, they largely rely on additional domain-specific model heads~\cite{zhu2023motionbert,zhang2025lmm} and require complex multi-stage training~\cite{ci2023unihcp,wu2024macdiff}, limiting their generalizability and scalability.

To overcome these limitations, we introduce a new \textbf{setting} that enables training a unified cross-domain model in a single process, eliminating the need for domain-specific components and complex multi-stage training. 
Motivated by in-context learning~\cite{brown2020language,liu2021makes} in natural language processing, which allows models to perform multiple tasks without explicit fine-tuning or retraining, we find that this paradigm aligns well with our proposed setting. 
While in-context learning has been extended to image-based tasks~\cite{zhang2024makes,wang2023painter} and point cloud-based tasks~\cite{fang2023pic}, its application to 3D human motion modeling remains unexplored, to the best of our knowledge.

\begin{figure*}[tp]
    \hsize=\textwidth
    \centering
    \includegraphics[width=0.99\textwidth]{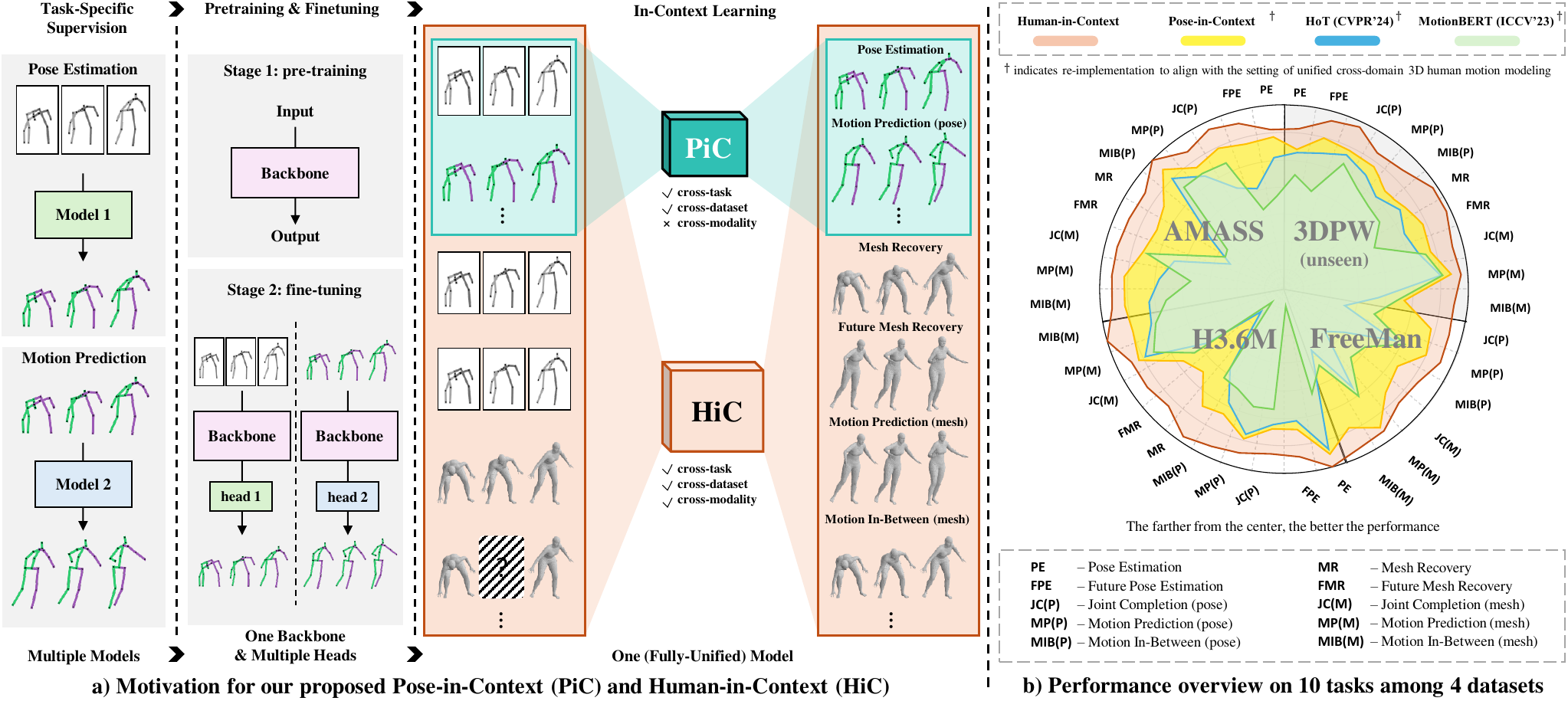}
    \vspace{-0.5em}
    \caption{
    Motivation (left) and performance overview (right) for our proposed Pose-in-Context and Human-in-Context. Compared to previous methods, which require multiple models/heads along with multiple training stages, PiC and HiC unify various human-centric tasks and datasets through one-time training with a fully unified model. Furthermore, HiC extends PiC into a more inclusive, scalable, and generalizable cross-domain model by introducing designs in both functionality and technicality.
    }
    \label{fig:teaser}
    \vspace{-1em}
\end{figure*}

We present PiC, the first approach that incorporates in-context learning into 3D human motion modeling to facilitate a pose-centric cross-domain model. 
PiC achieves generalization across various pose-based tasks and datasets, showing competitive performance. 
However, PiC has three key limitations:
1) PiC focuses on the scope of pose-based tasks and datasets without the ability to generalize across different modalities;
2) PiC uses a random selection-based prompting strategy, selecting prompts independently of queries, which could result in a large gap between the query and the selected prompt; and
3) PiC employs an attention-based network architecture, which does not exploit contextual dependencies other than global context.

To overcome the limitations of PiC, we develop Human-in-Context, a cross-domain model in 3D human motion modeling that achieves cross-modality, cross-task, and cross-dataset generalization within a single unified framework.
Unlike PiC, which addresses only pose-based tasks, HiC can handle both pose- and mesh-based tasks through one-time training.  
HiC extends PiC in three aspects (see Figure \ref{fig:teaser}).
First, HiC extends the model’s generalization capability to cover a broader range of domains, including both pose and mesh modalities. 
By representing pose- and mesh-based human motion within a unified formulation, HiC supports cross-modal learning within a single in-context framework. 
Regarding task coverage, for each pose-based task in PiC, HiC includes a corresponding mesh-based variant, effectively doubling the number of supported tasks. 
Second, HiC employs a max-min similarity prompt sampling strategy to enhance generalization. Instead of random selection, it samples representative anchors from the training set, retrieving the closest-matching anchor for each query to ensure contextual alignment. This approach dynamically pairs hard anchors (fixed representatives) with soft anchors (adaptable refinements), improving generalization across diverse domains (both in-distribution and out-of-distribution).
Third, HiC introduces a new network, \model, with a dual-branch architecture, enabling context integration across prompts and queries. 
Each branch consists of a sequence of \modelname~blocks to handle contextual information at multiple levels. 
Each block performs two operations: multi-level context aggregation and cross-level context update. 
In the aggregation step, features are processed using state-space models, self-attention, and graph convolution to capture dependencies across different views, scopes, and feature spaces. 
In the update step, the network adjusts feature representations across levels by estimating their relative importance for the target task. 
These operations support context-aware representation learning across diverse motion domains.

Extensive results show that PiC and HiC perform favorably against state-of-the-art domain-specific and cross-domain approaches. 
Figure \ref{fig:teaser} shows the overall framework where PiC\footnote{Pose-in-Context is referred to as ``Skeleton-in-Context'' in our prior work. 
For terminological consistency in the cross-domain context of this journal version, we use ``pose'' instead of ``skeleton'' throughout this paper.} is first proposed in our \textit{CVPR 2024} conference paper~\cite{wang2024sic}. This work, HiC, extends PiC in several aspects:

\begingroup
\setlist[itemize,1]{leftmargin=0.5cm}
\begin{itemize}
    \item
    HiC is an in-context cross-domain model in 3D human motion modeling that simultaneously possesses cross-modality, cross-task, and cross-dataset generalization capability.
    \item 
    HiC incorporates a larger scope of domains, including two modalities, ten tasks, and four datasets, while expanding the scale of data by $\sim21\times$.
    \item 
    HiC proposes a prompting strategy, max-min similarity prompt sampling, to tackle the challenge of generalizing across highly diverse domains.
    \item
    HiC proposes a network, \model, to effectively capture complex in-context dependencies through multi-level context aggregation and cross-level context update.
    \item 
    Extensive experimental results show that HiC consistently delivers superior performance in both in-domain and out-of-domain generalization, outperforming PiC by 9.9\% and other domain-specific/cross-domain models by 21.8\% on average.
\end{itemize}
\endgroup

\section{Related Work}
\label{sec:related_work}

\noindent\textbf{3D Human Motion Modeling.}
3D human motion modeling covers a broad variety of domains, involving different tasks and modalities.
Some widely recognized tasks include motion prediction~\cite{martinez2017human,li2018convolutional,li2021multiscale,mao2019learning,shu2021spatiotemporal}, pose estimation\cite{martinez2017simple,zhang2022mixste,li2022mhformer,gong2023diffpose,zhang2024app,xu2024finepose}, mesh recovery~\cite{choi2020pose2mesh,tang2024arts,li2021hybrik,yu2021skeleton2mesh}, and motion in-between~\cite{yan2019convolutional,zhou2020generative,kaufmann2020convolutional,hernandez2019human}, to name a few.
Regarding modalities, poses~\cite{yao2012coupled,gong2023diffpose,liu2019feature} and meshes~\cite{SMPL:2015,bogo2016keep} represent two of the most popular ways to represent 3D human motion.
Compared to 2D representations such as RGB images/videos~\cite{li2018unsupervised,wang2020motion}, 3D representations offer a more efficient, compact, and informative way of modeling human motion.
Regarding 3D human motion datasets, Human3.6M~\cite{ionescu2013h36m}, one of the most commonly used datasets for 3D human motion modeling, offers high-quality 3D joint positions captured from a large number of subjects performing diverse activities. 3DPW~\cite{von2018_3dpw} is designed for outdoor environments, featuring real-world motion and complex camera perspectives. AMASS~\cite{mahmood2019amass} provides a collection of motion capture data from multiple sources (such as CMU-Mocap) using SMPL~\cite{SMPL:2015} parametrization. FreeMan~\cite{wang2024freeman} is a more recent dataset focused on both pose-based and mesh-based human motion, designed to improve the realism and diversity of motion data for a variety of tasks in 3D human motion modeling.
NTU RGB+D~\cite{shahroudy2016ntu60} and NTU RGB+D 120~\cite{liu2019ntu} are used mainly for classification tasks~\cite{liu2020disentangling,liu2017enhanced,vg4dICRA24}, rather than tasks that output motion sequences.
Different domains within 3D human motion modeling could have significant gaps, making existing works dependent on domain-specific designs when addressing multiple domains.
In contrast, we introduce PiC and HiC, achieving cross-domain 3D human motion modeling without any domain-specific designs.

\noindent\textbf{Human-Centric Domain-Specific Models.}
Early works address individual domains by designing domain-specific models~\cite{chen20173d,liu2020comprehensive}.
Numerous methods design domain-specific models include motion prediction using 3D pose data from Human3.6M~\cite{mao2019learning,ma2022progressively,xu2021graph} or AMASS dataset~\cite{mao2020history}, mesh recovery using 3DPW dataset~\cite{choi2020pose2mesh,li2021hybrik}, etc.
In recent years, various backbones have been proposed for domain-specific models, including MLP~\cite{guo2023back}, convolutional~\cite{yan2019convolutional,kaufmann2020convolutional,li2018convolutional,pavllo20193d}, recurrent~\cite{zhu2016co,liu2016spatio,martinez2017human},  graph~\cite{choi2020pose2mesh,li2021multiscale,wang2024dynamic,cai2019exploiting,cui2020learning} neural networks, and transformers~\cite{zheng20213dposetrans,li2024hot,li2022mhformer,zhao2022graformer,mehraban2024motionagformer}, which are trained individually on different datasets such as Human3.6M and 3DPW.
Additionally, Mamba~\cite{gu2023mamba,zhu2024vision} has been used for 3D human motion modeling~\cite{zhang2024pose,chaudhuri2024simba}.
%
%
In contrast to domain-specific models, which require deploying standalone models and training them separately in individual domains, we leverage in-context learning to establish a unified approach to address multiple domains more efficiently and effectively.

\noindent\textbf{Human-Centric Cross-Domain Models.}
In contrast to domain-specific models, recent approaches focus on unifying different domains by designing cross-domain models~\cite{li2024omg,devlin2018bert,wang2023seggpt,wang2023painter,fang2023pic}.
Due to multidimensional and spatial-temporal signals of human motion, existing cross-domain models are limited to a small number of domains while still requiring additional domain-specific model components or multi-stage training.
For example, motion prediction models trained on AMASS and tested on 3DPW~\cite{mao2020history,guo2023back}, or mesh recovery models are trained on Human3.6M and tested on 3DPW~\cite{tang2024arts}.
Cross-domain modeling also involves generalizing across several tasks, e.g., pose and shape estimation~\cite{kolotouros2019spin,bogo2016keep}, pose estimation and action recognition~\cite{luvizon2018poseaction,yao2012coupled}, and motion prediction and action recognition~\cite{li2021symbiotic}.
Recent cross-domain models can generalize across more tasks on single-modal data~\cite{cui2021towards,cai2021unified}.
With new backbones~\cite{dosovitskiy2020image} and training paradigms~\cite{devlin2018bert,brown2020language} emerging, cross-domain models have demonstrated promising results.
UniHCP~\cite{ci2023unihcp} performs several human perception tasks by using a shared backbone paired with a task-aware head to switch between tasks, which is trained under task-specific supervision.
In~\cite{zhu2023motionbert}, MotionBERT pre-trains a backbone to learn a unified representation and fine-tunes different task heads individually to fit different tasks.
On the other hand, UPS~\cite{foo2023UPS} can perform three different tasks by using task-specific embeddings to activate different blocks in the model and regressively generate tokens in a language-like manner.
LMM~\cite{zhang2025lmm} generalizes over several tasks by fine-tuning dataset-specific layers on top of a unified pre-trained representation with task-specific conditions as additional inputs required for different tasks.
Nevertheless, the ineffective backbone and the unavoidable involvement of domain-specific model heads or fine-tuning are two key limitations of existing cross-domain models.
To address these limitations, we propose a unified cross-domain model that can be trained once for all tasks.
To our knowledge, we are the first to design a model for various 3D human tasks.

\begin{table}[t]
    \centering
    \caption{Comparison between HiC and PiC.
    }
    \vspace{-1em}
    \renewcommand{\arraystretch}{1.2}
    \resizebox{\columnwidth}{!}{
    \begin{tabular}{l|r|ccc}\hline
        \multirow{2}{*}{\textbf{Name}}  & \multirow{2}{*}{\textbf{Data Scale}}  & \multicolumn{3}{c}{\textbf{Domain Scope}}  \\
                                        &                                       &     modality   & task & dataset             \\ \hline 
        Pose-in-Context             & 0.18M                                  &   pose     &   5  &    3                \\
        Human-in-Context                & 3.83M                                & pose \& mesh  &   10  &    4              \\ \hline
    \end{tabular}
    }
    \renewcommand{\arraystretch}{1}
    \label{tab:sic hic compare}
    \vspace{-1em}
\end{table}

\begin{table*}[t]
    \centering
    \caption{Human-in-Context setups. Within the scope of this work, a domain is interpreted in three aspects, including the input and output of the task, the modalities it involves, and the applicable datasets.
    }
    \vspace{-0.5em}
    \renewcommand{\arraystretch}{1.55}
    \resizebox{\textwidth}{!}{
    \begin{tabular}{@{}m{0.2cm}@{\hspace{0.25cm}}m{3cm}@{\hspace{0.2cm}}m{1.2cm}@{}|m{1.95cm}m{1.65cm}|m{0.25cm}<{\centering}m{0.25cm}<{\centering}m{0.45cm}<{\centering}|m{0.5cm}<{\centering}m{0.7cm}<{\centering}m{0.7cm}<{\centering}m{0.7cm}<{\centering}} \hline
        & \multirow{2}{*}{\textbf{Domain}} && \multicolumn{2}{c|}{\textbf{Task}} & \multicolumn{3}{c|}{\textbf{Modality}} & \multicolumn{4}{c}{\textbf{Dataset}} \\
        &&& \multicolumn{1}{c}{Input} & \multicolumn{1}{c|}{Output} & \shortstack[c]{pose\\(2D)} & \shortstack[c]{pose\\(3D)} & mesh & H3.6M & AMASS & FreeMan & 3DPW \\ \hline 
        1.& Pose Estimation &PE& $\mathbf{X}^{\text{2D\_pose}}_{1:F}$ & $\mathbf{X}^{\text{3D\_pose}}_{1:F}$ 
            & \checkmark & \checkmark &            & \checkmark & \checkmark &            & \checkmark \\ 
        2.& Future Pose Estimation &FPE& $\mathbf{X}^{\text{2D\_pose}}_{1:F}$ & $\mathbf{X}^{\text{3D\_pose}}_{1:F}$ 
            & \checkmark & \checkmark &            & \checkmark & \checkmark &            & \checkmark \\ 
        3.& Mesh Recovery &MR& $\mathbf{X}^{\text{2D\_pose}}_{1:F}$ & $\{ \mathbf{X}^{\text{mesh}}_{1:F}, \bm{\beta} \}$
            & \checkmark &            & \checkmark & \checkmark & \checkmark &            & \checkmark \\ 
        4.& Future Mesh Recovery &FMR& $\mathbf{X}^{\text{2D\_pose}}_{1:F}$ & $\{ \mathbf{X}^{\text{mesh}}_{F+1:2F}, \bm{\beta} \}$
            & \checkmark &            & \checkmark & \checkmark & \checkmark &            & \checkmark \\ 
        5.& Motion Prediction (pose) &MP (P)& $\mathbf{X}^{\text{3D\_pose}}_{1:F}$ & $\mathbf{X}^{\text{3D\_pose}}_{F+1:2F}$ 
            &            & \checkmark &            & \checkmark & \checkmark & \checkmark & \checkmark \\ 
        6.& Motion In-Between (pose) &MIB (P)& $\mathbf{X}^{\text{3D\_pose}}_{1:F} \odot \mathbf{M}^{\text{time}}$ & $\mathbf{X}^{\text{3D\_pose}}_{1:F}$
            &            & \checkmark &            & \checkmark & \checkmark & \checkmark & \checkmark \\ 
        7.& Joint Completion (pose) &JC (P)& $\mathbf{X}^{\text{3D\_pose}}_{1:F} \odot \mathbf{M}^{\text{joint}}$ & $\mathbf{X}^{\text{3D\_pose}}_{1:F}$
            &            & \checkmark &            & \checkmark & \checkmark & \checkmark & \checkmark \\ 
        8.& Motion Prediction (mesh) &MP (M)& $\mathbf{X}^{\text{mesh}}_{1:F}$ & $\{ \mathbf{X}^{\text{mesh}}_{F+1:2F}, \bm{\beta} \}$
            &            &            & \checkmark & \checkmark & \checkmark & \checkmark & \checkmark \\ 
        9.& Motion In-Between (mesh) &MIB (M)& $\mathbf{X}^{\text{mesh}}_{1:F} \odot \mathbf{M}^{\text{time}}$ & $\{ \mathbf{X}^{\text{mesh}}_{1:F}, \bm{\beta} \}$
            &            &            & \checkmark & \checkmark & \checkmark & \checkmark & \checkmark \\ 
        10.& Joint Completion (mesh) &JC (M)& $\mathbf{X}^{\text{mesh}}_{1:F} \odot \mathbf{M}^{\text{joint}}$ & $\{ \mathbf{X}^{\text{mesh}}_{1:F}, \bm{\beta} \}$
            &            &            & \checkmark & \checkmark & \checkmark & \checkmark & \checkmark \\ \hline
    \end{tabular}
    }
    \renewcommand{\arraystretch}{1}
    \label{tab:tasks}
    \par
    \textit{\footnotesize 
    Some conventional task names, such as motion prediction, may lack specificity in the modality; in such cases, the modality is indicated in parentheses.}
\end{table*}

\noindent\textbf{In-Context Learning.}
In-context learning~\cite{brown2020language,rubin2021learning,radford2021learning} presents a novel approach to performing multiple tasks based on context, without explicit task-specific retraining or fine-tuning, which has been recently been used in vision and natural language processing~\cite{wang2023promptdiffusion,bar2022visualprompt,fang2023pic,wang2024refldmseg}.
A context is presented in the form of a prompt~\cite{wu2023towards,rubin2021learning,xu2024finepose}, which consists of a pair of input and target ouput, serving as an example of what the task is expected to accomplish~\cite{zhang2024makes,liu2021makes}.
In-context learning significantly relies on two key design elements: prompting strategy~\cite{sun2023exploring} and network architecture~\cite{rubin2021learning}.
An effective prompting strategy ensures that prompts are designed and employed to provide sufficient context for the model to learn from.
Meanwhile, a well-designed network architecture ensures that the model effectively processes the prompts to extract the hidden task from the context and then performs the desired task on the queries.
While in-context learning aligns well with the setting of unified cross-domain 3D human motion modeling, its application to human motion remains unexplored.
In this work, we introduce in-context learning into 3D human motion modeling, achieving a unified cross-domain model with strong scalability and generalization.
\section{Problem Formulation}
\label{sec:problem formulation}

In this section, we introduce basic concepts necessary for Human-in-Context, including the cross-domain setup and in-context learning.

\subsection{Human-in-Context Setups}
\label{sub_sec:cross domain setup}
Unifying different domains is a crucial preliminary step in building a cross-domain model. In this work, domain refers to the combination of a specific task, the modality(ies) it involves, and the dataset(s) it applies to. The domains addressed within the scope of Human-in-Context are shown in detail in Table~\ref{tab:tasks}.

\noindent\textbf{Cross-Modal Setup.}
    To facilitate cross-modality generalization for Human-in-Context, we first re-interpret pose-based and mesh-based representations in a unified formulation, enabling them to be jointly applied in a cross-modal setting.
    Poses and meshes are two different modalities of human representation, which are not commonly studied jointly in a unified framework by existing works in 3D human motion modeling.
    Compared to the gaps between different tasks or datasets, the domain gap between different modalities usually presents a more significant challenge.

    \textbf{1)} Pose-based human motion is represented as a sequence of poses, where a pose consists of multiple joints represented by position coordinates in 2D or 3D space. Let a sequence of poses be $\mathbf{X}_{1:F}^{\text{pose}}=[ \mathbf{x}^{\text{pose}}_1, \mathbf{x}^{\text{pose}}_2, \cdots, \mathbf{x}^{\text{pose}}_F ] \in \mathbb{R}^{F\times N \times C}$, where $\mathbf{x}^{\text{pose}}_f$ is the pose at $f$-th frame consisting of $N$ joints. For the case of 2D pose $\mathbf{X}^{\text{2D\_pose}}$, the joint is represented by $C=2$ coordinates $(x,y)$, and for the case of 3D pose $\mathbf{X}^{\text{3D\_pose}}$, the joint is represented by $C=3$ coordinates $(x,y,z)$.
    To obtain a unified pose format before unifying it with the mesh format, we extend 2D poses into 3D by appending an all-zero slice along the z-axis.
    
    \textbf{2)} Mesh-based human motion is represented as a sequence of 3D meshes consisting of vertices and faces.
    This mesh-based approach captures not only the global body joint rotation configurations but also the finer geometric details of the body's surface. 
    A commonly-used parametrization of mesh representations is the Skinned Multi-Person Linear (SMPL)~\cite{SMPL:2015} model, which takes the joint rotation parameters $\bm{\theta}\in\mathbb{R}^{3J}$ and the shape parameters $\bm{\beta}\in\mathbb{R}^{S}$ as inputs and then outputs a triangulated mesh $\mathcal{V}\in\mathbb{R}^{6980\times 3}$.
    The SMPL joint rotation vector $\bm{\theta}$ consists of $J$ joints, each represented by a 3-dimensional axis-angle rotation vector $(\theta_x, \theta_y, \theta_z)$, where $(\theta_x, \theta_y, \theta_z)$ is the axis of rotation scaled by the angle of rotation in radians. 
    The SMPL shape vector $\bm{\beta}$ encodes the individual body shape variations, such as height and weight. Let a sequence of SMPL joint rotation vectors be $[ \bm{\theta}_1, \bm{\theta}_2, \cdots, \bm{\theta}_F ] \in \mathbb{R}^{F\times 3J}$, where $\bm{\theta}_f$ is the SMPL joint rotation vector at $f$-th frame. 
    To align the dimensions of $\bm{\theta}_f$ with the dimensions of $\mathbf{x}_{f}^{\text{pose}}$, we re-organize the elements of $\bm{\theta}_f$ to obtain $\mathbf{X}_{1:F}^{\text{mesh}}=[ \mathbf{x}^{\text{mesh}}_1, \mathbf{x}^{\text{mesh}}_2, \cdots, \mathbf{x}^{\text{mesh}}_F ] \in \mathbb{R}^{F\times J \times 3}$. 
    
    To unify cross-modal data among pose- and mesh-based representations, we first find the maximum joint count $max(N, J)$, where $N$ and $J$ are the joint numbers in $\mathbf{X}_{1:F}^{\text{pose}}$ and $\mathbf{X}_{1:F}^{\text{mesh}}$.
    We then append virtual joints, whose elements are all zeros, to data with fewer joints than $max(N, J)$. Additionally, for pose-based representations $\mathbf{X}_{1:F}^{\text{pose}}$, we assign virtual shape parameters $\bm{\beta}=\bm{0}$, which are only to align the data format of both pose- and mesh-based representations and will not affect training or evaluation.
    Without loss of generality, assume $J=max(N, J)$, a unified cross-modal human motion representation is defined as $\mathbf{X}_{1:F} = [ \mathbf{x}_1, \mathbf{x}_2, \cdots, \mathbf{x}_F ] \in \mathbb{R}^{F\times J \times C}$, where the elements can be position coordinates or axis-angle rotation vectors.

\begin{figure*}[tp]
    \hsize=\textwidth
    \centering
    \includegraphics[width=0.99\textwidth]{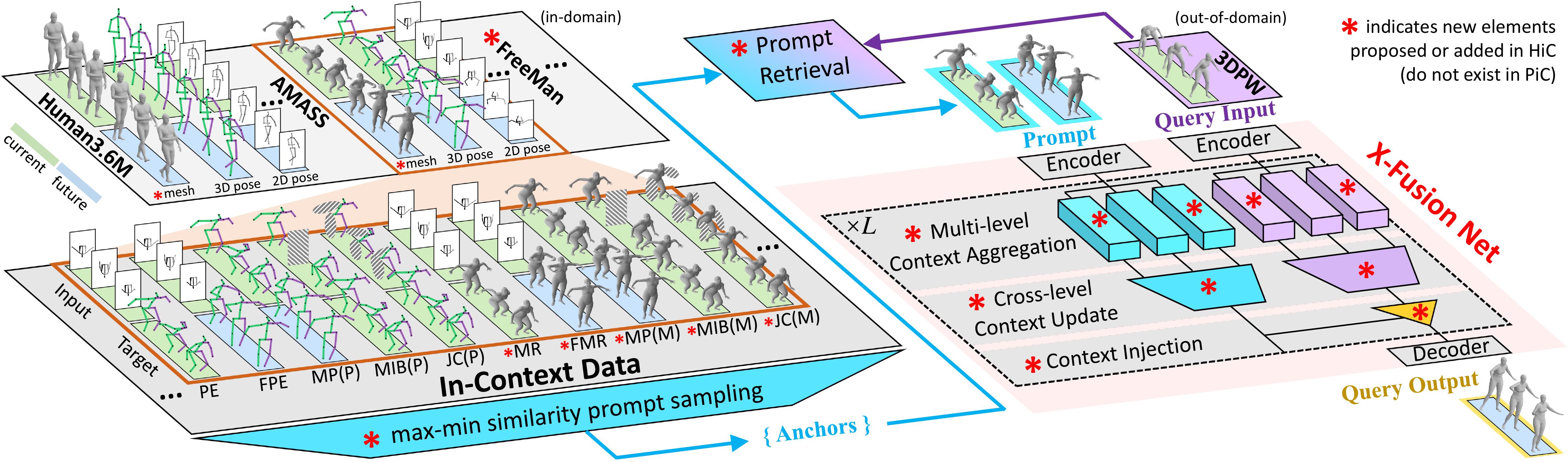}
    \vspace{-0.5em}
    \caption{
    Pipeline of the proposed Human-in-Context (HiC), with highlights on its difference from PiC. On the data end, from various datasets containing motion clips represented in multiple forms, we derive the in-context datasets containing input-target pairs for 10 different tasks.
    A collection of anchors is sampled from the in-context datasets using the proposed max-min similarity prompt sampling.
    In each inference, the most suitable prompt is retrieved from the anchors based on the query input (in this example an out-of-domain motion sequence from 3DPW dataset).
    On the network end, the query input and the prompt are fed through the proposed \model, where we apply multi-level context aggregation, cross-level context update, context injection, and other modules.
    }
    \label{fig:framework}
    \vspace{-1em}
\end{figure*}

\noindent\textbf{Cross-Task Setups.}
3D human motion tasks can be derived from any motion sequence $\mathbf{X}_{1:2F}\in\{ \mathbf{X}^{\text{2D\_pose}}_{1:2F}, \mathbf{X}^{\text{3D\_pose}}_{1:2F}, \mathbf{X}^{\text{mesh}}_{1:2F} \}$ by specifying inputs and targets on different modalities and/or in different time windows.
As $\mathbf{X}^{\text{2D\_pose}}_{1:2F}$, $\mathbf{X}^{\text{3D\_pose}}_{1:2F}$, and $\mathbf{X}^{\text{mesh}}_{1:2F}$ describe the same human motion clip from different perspectives, we can build a unified formulation for different tasks.
Specifically, we explore 10 tasks, involving different modalities and time windows, whose inputs and targets are shown in Table~\ref{tab:tasks}. 

Assume the first $F$ frames represent the ``current (or history)'', and the last $F$ frames represent the ``future''.
Pose estimation and mesh recovery tasks focus on estimating the current motion in another modality.
Tasks such as motion prediction aim to predict future motion in the same modality as historical motion.
On the other hand, future pose estimation and mesh recovery tasks are more challenging as they are required to predict future motion in a new modality.
Motion in-between and joint completion tasks are expected to recover missing parts of the motion. 

To formulate these tasks, we apply a binary mask $\mathbf{M}^{\text{time}}\in\mathbb{R}^{F}$ or $\mathbf{M}^{\text{joint}}\in\mathbb{R}^{J}$ to the input clip, which determines which specific frames or joints are missing. 
The masks are randomly generated for each motion clip and then padded to have the same shape as the motion clip.
To avoid changing the nature of the task, the first and last frames, and the root joint are never masked.
Regardless of the task definitions, their inputs and outputs all have the same shape and can fit in a single unified framework.

\noindent\textbf{Cross-Dataset Setups.}
    Training a unified cross-domain model requires sufficient data across large-scale datasets.
    However, these datasets do not all provide data in both poses and meshes. 
    Therefore, we follow standard practices in existing approaches to additionally process the datasets to obtain data across both poses and meshes.
    As datasets such as 3DPW~\cite{von2018_3dpw} and FreeMan~\cite{wang2024freeman} originally provide both pose and mesh data, they do not need to be additionally processed.
    For datasets that provide 2D and 3D pose data but not mesh data, such as Human3.6M~(H3.6M)~\cite{ionescu2013h36m}, the ground-truth SMPL data are generated by applying MoSh~\cite{loper2014mosh} to the sparse 3D MoCap marker data, as is done in existing work~\cite{kanazawa2018end,zhang2021pymaf}.
    Note that the 2D and 3D poses in these datasets, even when corresponding to the same motion clip, do not usually share the same values of $(x,y)$ coordinates. The 2D poses correspond to the 2D image space as they represent the joints of a human projected onto the pixel coordinates of an image, while the 3D poses are defined in the 3D world coordinate space.
    For datasets that provide SMPL mesh data but do not provide 2D/3D pose data, such as AMASS~\cite{mahmood2019amass}, we use a dataset-specific joint regressor pre-trained by~\cite{bogo2016keep} to regress the SMPL vertices to 3D pose joints, consistently with common practices~\cite{kolotouros2019spin,zhu2023motionbert,li2021hybrik,tang2024arts}.
    Then, we project the 3D poses orthogonally onto the xy-plane to derive 2D poses, following the standard approach~\cite{zhu2023motionbert}.


\subsection{In-Context Learning}
\label{sub_sec:in context learning}
    We use the form $[ <input>, <target> ]^{D}$ to denote the context corresponding to any specific domain $D$ in Table~\ref{tab:tasks}.
    In general, an in-context learning framework uses a model $\mathcal{M}(\cdot)$ to take a prompt ($P$) and a query ($Q$) as inputs, and generate the desired output:
    \begin{equation}\label{eq:general icl}
        \mathcal{M} \left( \begin{array}{r@{}l}
            [ &<input>_{P} , <target>_{P} ]^{D}, \\
            &<input>_{Q}
        \end{array} \right) \rightarrow \ <output>_{Q},
    \end{equation}
    where the prompt and the query are sampled from the same domain $D$. The prompt input and prompt target jointly provide the necessary context for the model to understand the task implied by the context and to perform the same task for the query input.

    In-context learning depends on two key design elements: prompting strategy and network architecture.
    The prompting strategy defines how prompts are obtained from domain $D$, while the network architecture defines how the model $\mathcal{M}(\cdot)$ processes the prompts and queries.
    An effective prompting strategy ensures that prompts are designed and employed in a way that provides sufficient context for the model to learn from, while a well-designed network architecture ensures that the model processes the prompts and queries in a way that effectively extracts the hidden task from the context provided by the prompts, and then accurately performs the desired task on the queries.
    In this work, we propose two approaches, Pose-in-Context and Human-in-Context, to effectively implement in-context learning to achieve unified cross-domain 3D human motion modeling, which will be respectively presented in the following two sections.

    For notational simplicity, the subscript indicating the frame indices, such as $1:F$, will be omitted in the following sections, as the frame indices can be inferred according to different domains.

\begin{figure*}[tp]
    \hsize=\textwidth
    \centering
    \includegraphics[width=0.99\textwidth]{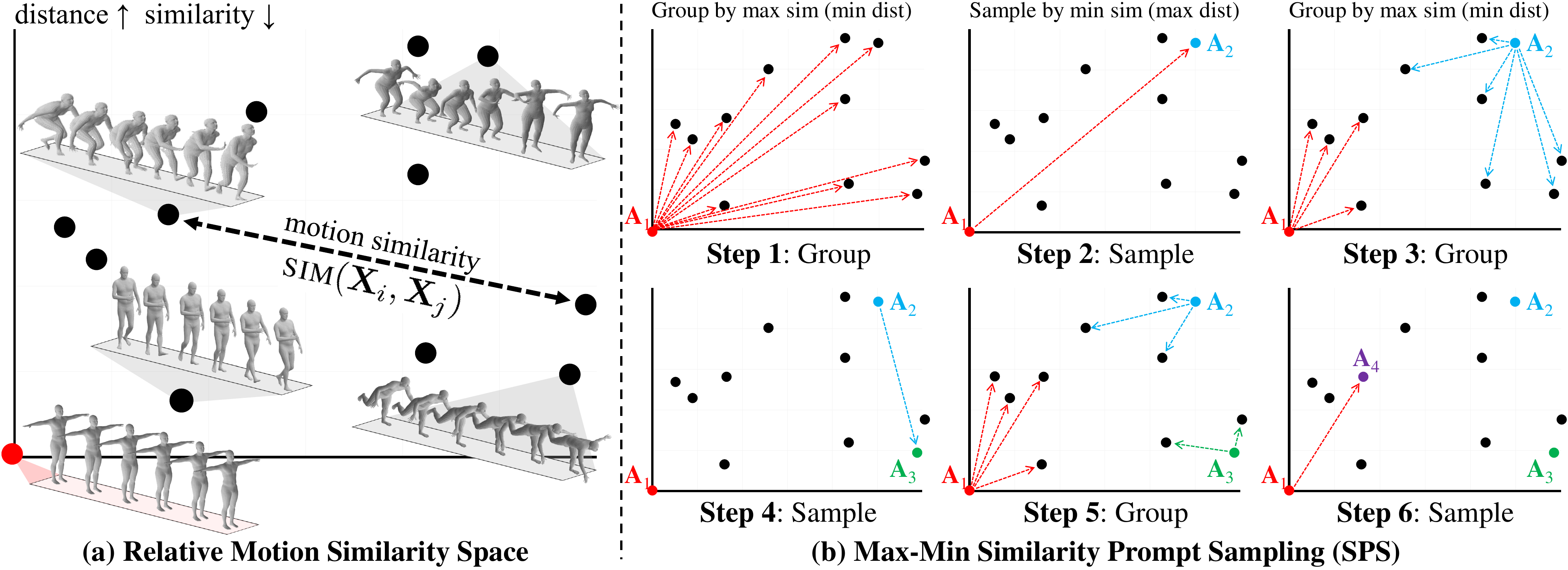}
    \vspace{-0.5em}
    \caption{
    Illustration of the prompting strategy in Human-in-Context.
    First, a relative motion similarity space is constructed on all motion sequences within the training scope, based on their relative similarity to a canonical body configuration and other motion sequences.
    Next, the max-min similarity prompt sampling operates iteratively to sample a set of anchors, which essentially involves two steps in each iteration: 1) grouping unsampled motion sequences by maximum similarity, and 2) sampling new anchors by minimum similarity.
    }
    \label{fig:prompt}
\end{figure*}

\section{Human-in-Context}
\label{sec:human-in-context}
We propose Human-in-Context, a cross-domain model that simultaneously possesses cross-modality, cross-task, and cross-dataset generalization capability, as shown in Figure~\ref{fig:framework}. 
Human-in-Context extends and improves Pose-in-Context in three aspects, including unlocking cross-modal generalization capability, expanding the scale of datasets and tasks, and improving performance across all domains.
The unlocked cross-modal generalization capability is enabled by the formulation of the cross-modal setup as explained in Section~\ref{sub_sec:cross domain setup}.
Furthermore, Human-in-Context introduces a prompting strategy and network architecture, respectively presented in Section~\ref{sub_sec:hic:prompting} and \ref{sub_sec:hic:net arch}, which jointly contribute to the improved performance across a larger scale of datasets and tasks in both poses and meshes.

\subsection{Pose-in-Context}

Pose-in-Context aims to generalize across various pose-based tasks and datasets, with a random selection-based prompting strategy and an attention-based network architecture implemented.
As Pose-in-Context is elaborated in our conference paper~\cite{wang2024sic}, we provide a brief review of its key designs in this subsection.

In PiC, the prompting strategy is based on a combination of Task-Guided Prompt (TGP) and Task-Unified Prompt (TUP). 
Consistently with the notations introduced in Section~\ref{sec:problem formulation}, a human motion sequence is denoted as $\mathbf{X} \in \mathbb{R}^{F\times J \times C}$, consisting of $F$ frames, $J$ joints, and $C$ channels.
TGPs are essentially randomly selected hard prompts, which can be written as $[ \mathbf{X}_{i}^\text{in}, \mathbf{X}_{i}^\text{gt} ]^{D}$, consistent with Eq.~(\ref{eq:general icl}), where $D$ represents the domain from which the TGP is randomly selected, and $i$ denotes the $i$-th sample in domain $D$. 
TGP aims to provide the necessary context information for the model to perform the task.
To further strengthen the in-context capability, we design an additional type of prompt, TUP, which adaptively learns to incorporate multiple tasks into a unified framework.
TUP is essentially a learnable soft prompt shared among all TGPs.
We introduce two approaches to implement TUP, denoted respectively as $\bar{\mathbf{U}}$ and $\widetilde{\mathbf{U}}$. 
This first approach uses a pseudo pose that contains prior domain information and then uses an encoder to project it into feature space. The pseudo pose is an average of all training poses.
Another approach applies the TUP directly in feature space, where TUP is factorized as the multiplication of two learnable prompts.

Given a query input $\mathbf{Q}_{j}^{D}$, which represents the $j$-th sample from domain $D$, a TGP $[ \mathbf{X}_{i}^\text{in}, \mathbf{X}_{i}^\text{gt} ]^{D}$ is randomly selected from the same domain $D$, representing the $i$-th sample.
After obtaining the query and the prompts based on the introduced prompting strategy, we map them into high-dimensional feature space using a shared encoder, obtaining a prompt feature $\mathbf{X}_P$ and a query feature $\mathbf{X}_Q$:
\begin{equation}\label{eq:sic input}
    \mathbf{X}_P = \mathcal{E}([\mathbf{X}_i^\text{in} , \mathbf{X}_i^\text{gt}]^{D});
    \mathbf{X}_Q = [\mathcal{E}(\mathbf{Q}_j^{D}) , \mathbf{U}],
\end{equation}
where $\mathcal{E}(\cdot)$ represents the encoder, $[\cdot,\cdot]$ denotes concatenation along the temporal axis, and the TUP $\mathbf{U}\in\{\bar{\mathbf{U}},\widetilde{\mathbf{U}}\}$ is agnostic to domain $D$ and can be implemented using either approach.
Motivated by \cite{zhu2023motionbert}, we implement the model $\mathcal{M}(\cdot)$ in Eq.~(\ref{eq:general icl}) using a two-steam spatial-temporal transformer, where the block in each stream consists of two branches, each applying spatial and temporal attention alternately in a different order. 
The prompt feature $\mathbf{X}_P$ and the query feature $\mathbf{X}_Q$ obtained from Eq.~(\ref{eq:sic input}) are the initial inputs to the query stream and the prompt stream, respectively. 
As the two streams share an identical structure, we take the prompt stream 
$l$-th layer as an example:
\begin{equation}\label{eq:sic model}
    \mathbf{X}_P^{l+1} = \alpha \mathcal{T}_{1} \big( \mathcal{S}_{1}\left ( \mathbf{X}_{P}^{l}\right ) \big) + \beta \mathcal{S}_{2}\ \big( \mathcal{T}_{2}\left ( \mathbf{X}_{P}^{l}\right ) \big),
\end{equation}
where $\alpha$ and $\beta$ are learnable balancing parameters, $\mathcal{S}$ and $\mathcal{T}$ respectively denote spatial and temporal attentions.
After obtaining $\mathbf{X}_P^{L+1}$ and $\mathbf{X}_Q^{L+1}$ through $L$ layers, they are mixed via an aggregation function $\mathcal{A}\left( \cdot\right)$ as $\mathbf{X}^{L+1}=\mathcal{A} ( \mathbf{X}_P^{L+1}, \mathbf{X}_Q^{L+1} )$, where $\mathcal{A}\left( \cdot\right)$ is a sum function with adaptive weights. 
Then, the model merges two streams into one, where the mixed feature $\mathbf{X}^{L+1}$ is fed through another $L'$ layers defined the same way as Eq.~(\ref{eq:sic model}).

\subsection{Prompting Strategy}
\label{sub_sec:hic:prompting}

In contrast to the random selection-based prompting strategy in PiC, which does not reflect any domain distribution patterns, we propose a max-min similarity prompt sampling (SPS) method in HiC, which introduces anchors to encode prior knowledge from the training data.
As illustrated in Figure~\ref{fig:prompt}, we obtain a compact collection of hard anchors, where each anchor encodes a cluster of context information from similar distributions. 
Each hard anchor is paired with a unique learnable soft anchor to dynamically adapt and refine the context.
When a query is presented to the model, it is compared to the hard anchors, and the most relevant hard anchor and its corresponding soft anchor are retrieved as the prompt, providing context information that reflects the cross-domain distribution.
This prompting strategy improves context-sensitive learning by incorporating the distributional structure of training data into the prompt construction process.

\noindent\textbf{Relative Motion Similarity Space.}
Using notations in Section~\ref{sec:problem formulation}, let $\mathbf{X}\in\mathbb{R}^{F\times J\times C}$ denote a motion sequence consisting of $F$ frames, $J$ joints, and $C$ channels, where the elements can represent position coordinates in poses or axis-angle rotations in meshes. Frame indices are omitted for notional simplicity.
We first define the similarity for any two arbitrary motion sequences $\mathbf{X}_{i}^{D_1}$ and $\mathbf{X}_{k}^{D_2}$, where $\mathbf{X}_{i}^{D_1}$ is the $i$-th sample in domain $D_1$ and $\mathbf{X}_{k}^{D_2}$ is the $k$-th sample in domain $D_2$. 
The similarity $\textsc{sim} (\cdot,\cdot)$ is defined as the average distance between two motion sequences:
\begin{equation}\label{eq:sim}
    \textsc{sim} (\mathbf{X}_i^{D_1}, \mathbf{X}_k^{D_2}) = - \frac{1}{FJ}\sum_{f=1}^F\sum_{j=1}^J \sqrt{\sum_{c=1}^C  \Big[ \mathbf{X}_i^{D_1} - \mathbf{X}_k^{D_2} \Big]_{(f,j,c)}^{2}},
\end{equation}
where $(f,j,c)$ indicates frame $f$, joint $j$, and channel $c$, and the negative sign in the equation is to ensure that smaller values indicate lower similarity (less similar), while larger values indicate higher similarity (more similar). 
This similarity measure reflects the average joint-level distance between two motion sequences.

Next, we construct a relative motion similarity space on the entire training set of motion sequences, which is defined as a feature space where each motion sequence is positioned based on its relative similarity to a canonical body configuration and its relationships to other motion sequences within the dataset. 
The canonical body configuration (T-body) is obtained by setting the joint rotation parameters of the SMPL~\cite{SMPL:2015} model as zeros, resulting in a human body naturally stretched like the letter T.
The canonical body configuration (T-body) serves as the reference point in this space.
Figure~\ref{fig:prompt}(a) shows a toy example of a relative motion similarity space constructed from 11 motion sequences, where each sequence is mapped to a point in a 2D space for illustrative purposes.
The relative motion similarity space facilitates understanding the distributive patterns of motion sequences and identify representative human motion from various domains.

\noindent\textbf{Max-Min Similarity Prompt Sampling (SPS).}
As introduced above, we propose the max-min similarity prompt sampling to iteratively select hard anchors within the constructed relative motion similarity space. 
Figure~\ref{fig:prompt}(b) shows an illustrative toy example of sampling 4 anchors from a set of 11 motion sequences step-by-step.
Given the entire training set of motion sequences $\mathcal{X} = \{\mathbf{X}_i^D\}$ for any sample $i$ in any domain $D$, we propose max-min similarity prompt sampling to sample a smaller set of $K$ motion sequences as hard anchors $\mathcal{A} = \{\mathbf{A}_k\}_{k=1}^{K}$.

During the max-min similarity prompt sampling, we keep track of two sets: $\mathcal{A}$ and $\mathcal{U}=\mathcal{X}\setminus\mathcal{A}$, representing the sets of the sampled and unsampled motion sequences, respectively.
To initialize the sampling process, we use the canonical body configuration (T-body) as the first anchor $\mathbf{A}_1$, providing a universal reference baseline.
We update the anchor set with the first anchor $\mathcal{A}=\{\mathbf{A}_1\}$, and remove it from $\mathcal{U}$.
The max-min similarity prompt sampling essentially involves iteration over two steps:

\textbf{1)} Group unsampled motion sequences by maximum similarity.
Using the similarity metric defined in Eq.~(\ref{eq:sim}), we compute the similarity between each $\mathbf{X}_i \in \mathcal{U}$ and each $\mathbf{A}_k \in \mathcal{A}$. 
For each $\mathbf{X}_i \in \mathcal{U}$, we find the anchor in $\mathcal{A}$ that has maximum similarity value to $\mathbf{X}_i$:
\begin{equation}
    \textsc{MaxSim}(\mathbf{X}_i, \mathcal{A}) = \max_{\mathbf{A} \in \mathcal{A}} \textsc{sim}(\mathbf{X}_i, \mathbf{A}).
\end{equation}
This step can be interpreted as grouping unsampled sequences based on the anchor to which it has the maximum similarity value, where the anchor serves as the ``group representative''. 
It ensures that sequences within the same group share similar patterns.

\begin{algorithm}[tp]
    \caption{Max-Min Similarity Prompt Sampling (SPS)}
    \begin{algorithmic}[1]
    \STATE \textbf{Input:} A set of motion sequences $\mathcal{X}$, number of anchors $K$, similarity function $\textsc{sim}(\cdot, \cdot)$ as defined in Eq.~(\ref{eq:sim})
    \STATE \textbf{Output:} A set of anchors $\mathcal{A}=\{\mathbf{A}_k\}_{k=1}^{K}$
    \STATE \textbf{Initialize:}
    \STATE Define $\mathbf{A}_1$ as a sequence of duplicated canonical bodies.
    \STATE $\mathcal{A} \gets \{\mathbf{A}_1\}$ \hfill // Set of anchors
    \STATE $\mathcal{U} \gets \mathcal{X} \setminus \mathcal{A}$ \hfill // Set of unsampled sequences

    \FOR{$k = 2$ to $K$}
        \STATE // Step 1: Group Unsampled Sequences by Max Similarity
        \FORALL{$\mathbf{X}_i \in \mathcal{U}$}
            \STATE Compute max similarity between $\mathbf{X}_i$ and anchors in $\mathcal{A}$:
            \vspace{-0.3em}
            \[
            \textsc{MaxSim}(\mathbf{X}_i, \mathcal{A}) = \max_{\mathbf{A} \in \mathcal{A}} \textsc{sim}(\mathbf{X}_i, \mathbf{A})
            \]
            \vspace{-1em}
        \ENDFOR
        \STATE // Step 2: Sample New Anchor by Min Similarity
        \STATE Find $\mathbf{X}^* \in \mathcal{U}$ with minimum $\textsc{MaxSim}(\mathbf{X}_i, \mathcal{A})$:
        \vspace{-0.3em}
        \[
            \mathbf{X}^* = \arg\min_{\mathbf{X}_i \in \mathcal{U}} \textsc{MaxSim}(\mathbf{X}_i, \mathcal{A})
        \]
        \vspace{-1em}
        \STATE Define $\mathbf{A}_k$ as $\mathbf{X}^*$:
        \vspace{-0.2em}
        \[\mathbf{A}_k \gets \mathbf{X}^*\]
        \vspace{-2em}
        \STATE // Step 3: Update Sets
        \STATE $\mathcal{A} \gets \mathcal{A} \cup \{\mathbf{A}_k\}$ \hfill // Add $\mathbf{A}_k$ to $\mathcal{A}$
        \STATE $\mathcal{U} \gets \mathcal{U} \setminus \{\mathbf{A}_k\}$ \hfill // Remove $\mathbf{A}_k$ from $\mathcal{U}$
    
        \IF{$\mathcal{U} = \emptyset$} 
            \STATE \textbf{break}
        \ENDIF
    \ENDFOR
    
    \RETURN $\mathcal{A}$
    \end{algorithmic}
    \label{alg:sps}
\end{algorithm}

\textbf{2)} Sample new anchors by minimum similarity.
Based on the grouped motion sequences and their corresponding values of $\textsc{MaxSim}(\mathbf{X}_i, \mathcal{A})$, we find $\mathbf{X}^*\in\mathcal{U}$ that has the minimum value among all $\textsc{MaxSim}(\mathbf{X}_i, \mathcal{A})$:
\begin{equation}
    \mathbf{X}^* = \arg\min_{\mathbf{X}_i \in \mathcal{U}} \textsc{MaxSim}(\mathbf{X}_i, \mathcal{A}).
\end{equation}
The new anchor to be sampled is defined to be $\mathbf{X}^*$, and the sets $\mathcal{A},\mathcal{U}$ are also updated accordingly:
\begin{equation}
    \begin{aligned}
        & \mathbf{A}_k \gets \mathbf{X}^*; \\
        & \mathcal{A} \gets \mathcal{A} \cup \{\mathbf{A}_k\}; \\
        & \mathcal{U} \gets \mathcal{U} \setminus \{\mathbf{A}_k\}.
    \end{aligned}
\end{equation}
This step can be interpreted as finding the group of motion sequences that exhibits the most diverse patterns, where motion sequences within the group are farther apart from each other than motion sequences in other groups. Then the new anchor is determined as the motion sequence farthest away from the ``group representative'', i.e., the anchor it is grouped to.
This process ensures that the sampled motion sequences are spread out diversely, covering different regions of the relative motion similarity space, and that each new anchor is the most distinct from the existing set of anchors, thus increasing the diversity of the sampled data.

The max-min similarity prompt sampling procedure stops when one of the following conditions is satisfied:
\begin{equation}
    \begin{aligned}
        & \text{Condition 1: }\vert\mathcal{A}\vert = K; \\
        & \text{Condition 2: }\mathcal{U} = \emptyset,
    \end{aligned}
\end{equation}
where $K$ is the pre-determined number of anchors, a hyperparameter balancing representational capacity and computational efficiency.
Although a larger $K$ increases the representational capacity by involving more anchors, it also increases computational loads.
In the extreme case $\mathcal{A}=\mathcal{X}$, i.e., $\mathcal{U} = \emptyset$, where all training data are set as anchors, it results in over-fitting and poor generalization capability. 
Thus, the hyperparameter $K$ in practice is set to be much smaller than the training scale. 
Specific values are determined through the ablation study as shown in Table~\ref{tab:abla:anchor num}.
Algorithm~\ref{alg:sps} summarizes the max-min similarity prompt sampling process.

\begin{figure}[tp]
    \hsize=\columnwidth
    \centering
    \includegraphics[width=0.99\columnwidth]{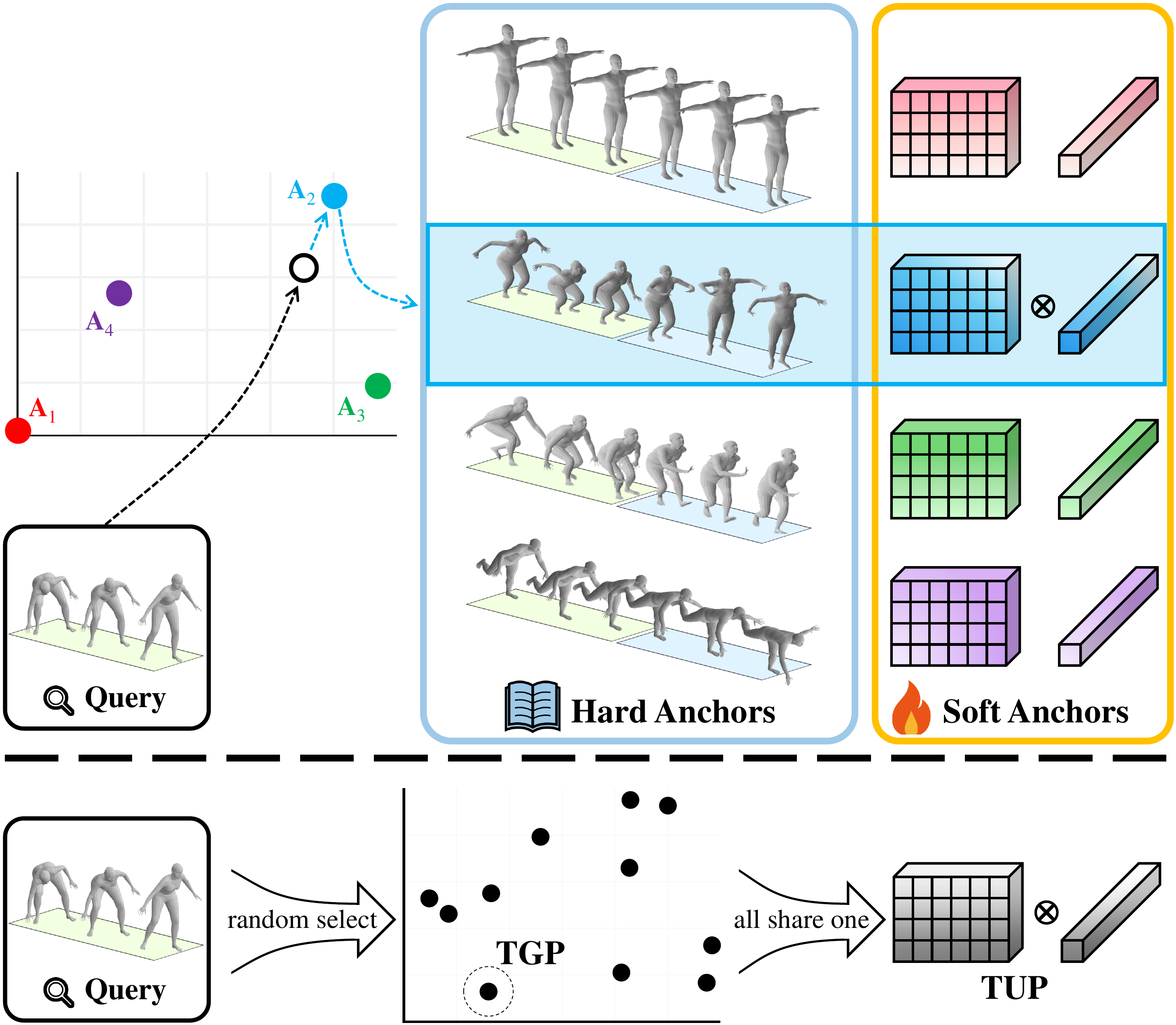}
    \vspace{-0.2em}
    \caption{
    Illustration of the prompt retrieval in HiC (top) and PiC (bottom). In HiC, a specific anchor is retrieved from the hard anchors based on their similarity to the query, and the corresponding soft anchor is also retrieved. In PiC, a TGP is selected randomly from all training data, and all TGPs share a single TUP.
    }
    \label{fig:prompt2}
    \vspace{-0.8em}
\end{figure}

\noindent\textbf{Hard and Soft Anchors.} 
Based on the set of hard anchors $\{ \mathbf{A}_k \}_{k=1}^{K}$ obtained by max-min similarity prompt sampling, we assign a soft anchor $\mathbf{U}_k$ to each $\mathbf{A}_k$. 
The soft anchor is defined in the high-dimensional hidden representation space as $\mathbf{U}_k = \mathbf{W}_1^k \mathbf{W}_2^k$, where $\mathbf{W}_1^k\in\mathbb{R}^{F\times J\times 1}$ and $\mathbf{W}_2^k\in\mathbb{R}^{1\times 1\times H}$ are trainable weights, and $H$ is the dimension of the hidden representation space.
As the hard anchors are sampled from a unified relative motion similarity space spanning various modalities, tasks, and datasets, they can extract domain-related information without requiring any domain-specific designs.
Soft anchors provide dynamic contextual complementary refinement, making the model more generalizable (see Table~\ref{tab:abla:tup}).
The use of hard and soft anchors jointly tackles the challenge of generalizing across a broad scope of diverse domains and a large scale of data, enabling the model to flexibly encode both domain-related information (via hard anchors) and generalizable domain patterns (via soft anchors) for improved adaptability.

\noindent\textbf{Prompt Retrieval.}
    As illustrated in Figure~\ref{fig:prompt2}, when a query is presented to the model, we retrieve the most relevant prompt from the anchors defined above in a similarity-based approach.
    First, we project the query input sequence $\mathbf{Q}^\text{in}$ into the same space as the hard anchors, i.e., relative motion similarity space.
    Second, we compute the similarity between $\mathbf{Q}^\text{in}$ and each hard anchor $\mathbf{A}_k$ using Eq.~(\ref{eq:sim}), and obtain the hard anchor with the highest similarity to the query input. These steps are described as follows:
    \begin{equation}
        \begin{aligned}
            \mathbf{A}^*& = \arg\max_{\mathbf{A}_k \in \mathcal{A}} \textsc{sim}(\mathbf{Q}^\text{in},\mathbf{A}_k ); \\
            \mathbf{P}^\text{in} &\leftarrow \mathbf{A}^*,
        \end{aligned}
    \end{equation}
    where $\mathbf{P}^\text{in}$ indicates prompt input. As the anchors originally come from the training set, we also obtain the ground-truth target $\mathbf{P}^\text{gt}$ corresponding to $\mathbf{P}^\text{in}$. 
    We then retrieve the soft anchor $\mathbf{U}^*$ corresponding to the hard anchor $\mathbf{A}^*$.
    The combination of prompt input $\mathbf{P}^\text{in}$ and prompt target $\mathbf{P}^\text{gt}$ represents a hard prompt, while $\mathbf{U}^*$ represents a soft prompt.
    The model takes $\mathbf{Q}^\text{in}$, $\mathbf{P}^\text{in}$, $\mathbf{P}^\text{gt}$, and $\mathbf{U}^*$ as inputs for further processing.
    The proposed prompting strategy contributes to Human-in-Context's ability to effectively adapt to different contexts involving various domains while maintaining robust generalization capabilities.

    In addition to the prompting strategy, another critical component of in-context learning is the network architecture, which defines how prompts and queries are processed to generate the final output in a specific context. 
    In Human-in-Context, we propose \model~to implement the network architecture.

\begin{figure*}[tp]
    \hsize=\textwidth
    \centering
    \includegraphics[width=0.99\textwidth]{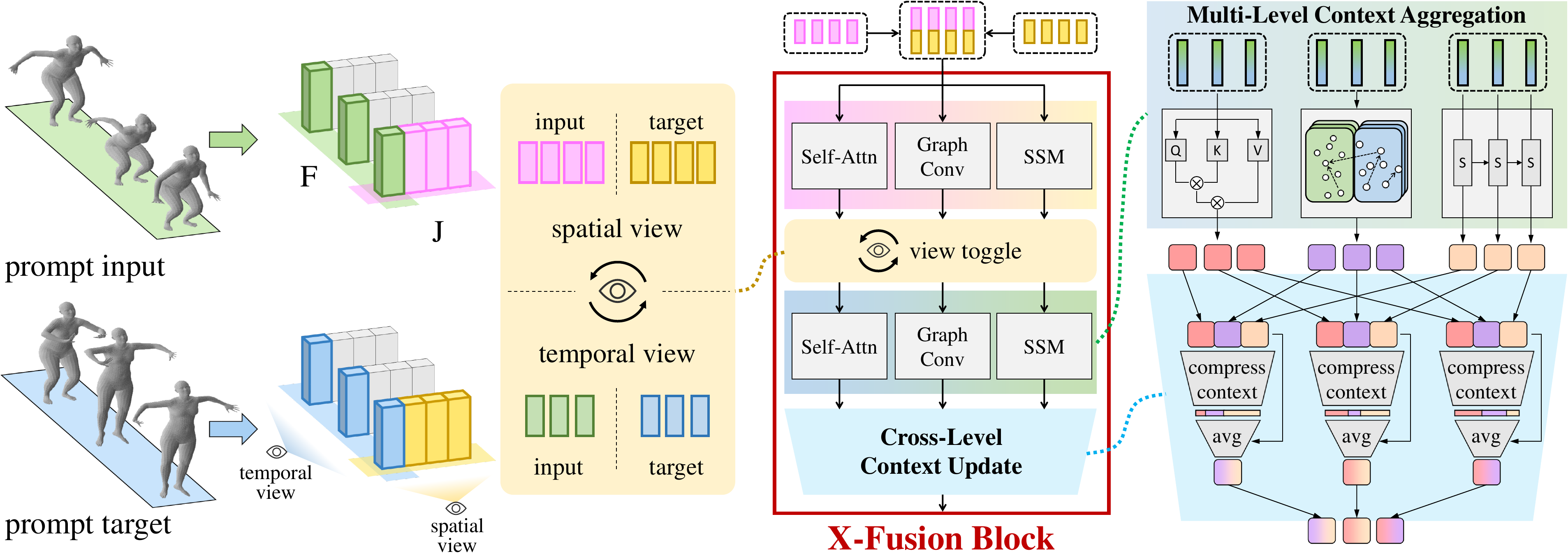}
    \vspace{-0.5em}
    \caption{
    Illustration of the \modelname~block. The \modelname~block applies multi-level context aggregation and cross-level context update to capture and fuse the contextual dependencies within and across different levels while toggling between the spatial and temporal views. 
    In Multi-Level Context Aggregation, we leverage self-attention, graph convolution, and state-space model to learn global dependencies in embedding space, local dependencies in graph space, and local dependencies in state-space, respectively.
    In Cross-Level Context Update, we compress context to remove redundancy and update features with emphasis on the most contributing levels.
    }
    \label{fig:network}
\end{figure*}

\subsection{\model}
\label{sub_sec:hic:net arch}
Given the exceptionally large scale of data spanning various modalities, tasks, and datasets that the network is expected to learn from simultaneously, a more effective and robust architecture than existing ones is required.
To this end, we propose \model, which is formulated as:
\begin{equation}
    \mathcal{M} ( \mathbf{Q}^\text{in},\mathbf{P}^\text{in},\mathbf{P}^\text{gt},\mathbf{U}^* ) \rightarrow \mathbf{Q}^\text{out},
\end{equation}
where $\mathbf{Q}^\text{in}, \mathbf{Q}^\text{out}, \mathbf{P}^\text{in}, \mathbf{P}^\text{gt}\in\mathbb{R}^{F\times J\times C}$ represent query input/output and prompt input/target, respectively, and $\mathbf{U}^*\in\mathbb{R}^{F\times J\times H}$ is the soft anchor, with $H$ denotes the dimension of hidden representations. 
Although $\mathcal{M}(\cdot)$ can be instantiated with any existing network that supports multiple inputs, the performance would not be satisfactory due to the great difficulty of in-context learning on multiple modalities, tasks, and datasets through one-time training. 
In Human-in-Context, we instantiate $\mathcal{M}(\cdot)$ by proposing \model, as shown in Figure~\ref{fig:framework}. \model~has a dual-branch structure, i.e., the query and prompt branches, with context injection between branches.
The query and prompt branches both employ \modelname~blocks as their basic blocks, whose architecture is shown in Figure \ref{fig:network}.

\subsubsection{Contextual Encoding}
    \model~is designed in a dual-branch structure consisting of the query ($Q$) and prompt ($P$), 
    \begin{equation}
        \begin{aligned}
            \mathbf{H}_{Q} &= \mathcal{E}_{Q} \big([\mathbf{Q}^\text{in} \parallel \mathbf{P}^\text{gt}]\big) + \mathbf{U}^*; \\
            \mathbf{H}_{P} &= \mathcal{E}_{P} \big([\mathbf{P}^\text{in} \parallel \mathbf{P}^\text{gt}]\big),
        \end{aligned}
    \end{equation}
    where $\mathbf{H}_{Q}$ and $\mathbf{H}_{P} \in \mathbb{R}^{F\times J\times H}$ represent the query and prompt contextual features, $[\cdot\Vert\cdot]$ denotes concatenation along the channel axis.
    In addition, $\mathcal{E}_{Q}$ and $\mathcal{E}_{P}$ are the query and prompt encoders, implemented with MLPs and positional embeddings~\cite{zhu2023motionbert}.
    The query and prompt contextual features are fed through the query and prompt branches. Each branch contains a sequence of \modelname~blocks as its basic modules.

\subsubsection{\modelname~Block}

Contextual representations entangle diverse dependencies that can be captured at different levels.
Specifically, contextual dependencies can be modeled in the spatial and temporal views, extracted from global and local scopes, and learned in embedding space, graph space, and state space.
To this end, the proposed \modelname~block contains two components: 1) multi-level context aggregation, and 2) cross-level context update. The two components jointly contribute to the ability to effectively capture and fuse the contextual dependencies within and across different levels.

Multilevel context aggregation is achieved through a set of aggregation functions, denoted as $\{\textsc{Aggregate}^{l}(\cdot)\}_{l=1}^{L}$ indexed by levels $l=1,2,\cdots,L$.
The cross-level context update is achieved through an update function, denoted $\textsc{Update}(\cdot)$.
Combining the multilevel aggregation functions and the update function, the \modelname~block is formulated as: 
\begin{equation}
    \begin{aligned}
        \mathbf{Z} &= \textsc{XFusion} (\mathbf{H})\\
        &= \textsc{Update}\big( \{\textsc{Aggregate}^{l}(\mathbf{H})\}_{l=1}^{L} \big),
    \end{aligned}
\end{equation}
where $\mathbf{H}\in\mathbb{R}^{F\times J\times H}$ denotes the input, and $\mathbf{Z}\in\mathbb{R}^{F\times J\times H'}$ denotes the output of the current \modelname~block.

Next, we discuss concrete implementations of these functions and explain their respective roles in the \modelname~block.

\vspace{0.5em}\noindent\textbf{Multi-Level Context Aggregation.}
Given contextual representations $\mathbf{H}\in\mathbb{R}^{F\times J\times H}$ from the previous layer, the aggregation functions $\{\textsc{Aggregate}^{l}(\cdot)\}_{l=1}^{L}$ at different levels $l$ are designed with the goals of extracting contextual features by focusing on different dependencies, whose respective outputs are:
\begin{equation}
    \mathbf{Y}^{l} = \textsc{Aggregate}^{l} (\mathbf{H}).
\end{equation}
First, we disentangle contextual dependencies in the spatial and temporal views. Regarding spatial-temporal dependencies, $\mathbf{H}\in\mathbb{R}^{F\times J\times H}$ can be modeled in spatial or temporal perspective.
In the spatial view, $\mathbf{H}$ is interpreted as spatial features per frame $\mathbf{H}_\text{S}\in\mathbb{R}^{J\times H}$, while in the temporal view, $\mathbf{H}$ is interpreted as per-joint temporal features $\mathbf{H}_\text{T}\in\mathbb{R}^{F\times H}$.
As the spatial and temporal features are processed similarly, we will use temporal features as an example and omit the subscripts S/T to demonstrate how multi-level context aggregation is applied.

Next, in each view, we apply aggregation functions at several levels using different implementations. Specifically, we employ self-attention (SelfAttn), graph convolution (GraphConv), and state-space model (SSM) to implement the aggregation functions as:
\begin{equation}\renewcommand\arraystretch{1.2}
    \begin{tabular}{ll}
    Implementation & Dependency \\\hline
    Level 1: $\mathbf{Y}^1 = \text{SelfAttn} (\mathbf{H})$ & embed-space; global \\
    Level 2: $\mathbf{Y}^2 = \text{GraphConv} (\mathbf{H})$ & graph-space; local \\ 
    Level 3: $\mathbf{Y}^3 = \text{SSM} (\mathbf{H})$ & state-space; local \\
    \end{tabular}
\end{equation}\renewcommand\arraystretch{1}
The collaborative strengths of self-attention, graph convolution, and the state-space model facilitate the network to capture dependencies reflected in different feature spaces and with different scopes, whose joint and individual contributions can be seen in Figure~\ref{fig:abla:netarch}.
During inference, multilevel aggregation is able to learn the most relevant dependencies dynamically for each prompt-query pair, as validated by the visualization in Figure~\ref{fig:viz mid feature}.

\vspace{0.5em}\noindent\textbf{Cross-Level Context Update.}
After obtaining a collection of features $\mathbf{Y}^l\in\mathbb{R}^{F\times H'}$ representing dependencies at various levels $l$ through multi-level context aggregation, we apply cross-level context update to generate the final output $\mathbf{Z}$ of the \modelname~block:
\begin{equation}
    \mathbf{Z} = \textsc{Update} \big( \{\mathbf{Y}^{l}\}_{l=1}^{L} \big).
\end{equation}
The cross-level context update is designed to update the aggregated features dynamically across different levels based on weighing their respective contributions in terms of generating the output accurately.
Specifically, the cross-level context update is applied frame-by-frame on $\{\mathbf{y}_{t}^{l}\}_{l=1}^{L}$ for each $t=1,2,\cdots,F$, where $\mathbf{y}_{t}^l\in\mathbb{R}^{H'}$ is the $t$-th row of $\mathbf{Y}^{l}$, meaning that the contributions of each level are evaluated frame-by-frame.

First, the influences of different levels at a specific frame $t$ are determined by compressing the contextual features $\{\mathbf{y}_{t}^{l}\}_{l=1}^{L}$ into a sequence of digits $\{a_{t}^{l}\}_{l=1}^{L}$, where each digit corresponds to a different level and represents its influence on the final output. The compression helps to remove redundancy and spot the most distinguishable information.
To begin the context compression step, the set of vectors $\{\mathbf{y}_{t}^{l}\}_{l=1}^{L}$ are concatenated along the channel dimension into a longer $LH'$-dimensional vector. The context compression step is defined as:
\begin{equation}
    \left[ \begin{array}{c} a_t^1 \\  a_t^2 \\  \vdots  \\ a_t^L  \end{array} \right]
    =
    \mathbf{W}^\text{compress} \left[ \renewcommand{\arraystretch}{1.2}\begin{array}{c}  \mathbf{y}_t^{1} \\\hdashline[2pt/2pt]  \mathbf{y}_t^{2} \\\hdashline[2pt/2pt]  \vdots \\\hdashline[2pt/2pt]  \mathbf{y}_t^{L} \\  \end{array} \right] + 
    \left[ \renewcommand{\arraystretch}{1}\begin{array}{c} b^1 \\  b^2 \\ \vdots \\ b^L  \end{array} \right],
\end{equation}
where \( \mathbf{W}^\text{compress} \in \mathbb{R}^{L \times LH'} \) is a trainable compression matrix shared across different frames $t$, and vector \( [b^1,b^2,\cdots,b^L]^\top \in \mathbb{R}^L \) is a trainable bias term associated with each aggregation level, also shared across different frames $t$. 
The vector obtained \( [a_t^1,a_t^2,\cdots,a_t^L]^\top \in \mathbb{R}^L \) contains scores that indicate the frame-specific influence of each level.
Then, we apply the softmax function to normalize the influence scores:
\begin{equation}
    [ \alpha_t^1, \alpha_t^2, \cdots, \alpha_t^L ]= \text{Softmax}( [a_t^1,a_t^2,\cdots,a_t^L] ).
\end{equation}
The final output $\mathbf{Z}\in\mathbb{R}^{F\times J\times H'}$ (in temporal view) of the cross-update feature update is obtained by fusing $\{\mathbf{y}_t^{l}\}_{l=1}^{L}$ according to the frame-specific influence scores $\alpha_t^l$. 
The frame-wise representation of $\mathbf{Z}$ is:
\begin{equation}
    \mathbf{z}_t = \sum_{l=1}^{L}  \alpha_t^l \mathbf{y}_t^l, 
\end{equation}
where $\mathbf{z}_t\in\mathbb{R}^{H'}$ is the $t$-th row in the final output $\mathbf{Z}$ of the current \modelname~block.
Based on the analogy between the temporal and spatial views, the joint representation $\mathbf{z}_j\in\mathbb{R}^{H'}$ of $\mathbf{Z}\in\mathbb{R}^{F\times J\times H'}$ (in the spatial view) can be inferred by $\mathbf{z}_j = \sum_{l=1}^{L}  \alpha_j^l \mathbf{y}_j^l$,
where $\{\mathbf{y}_j^l\}_{j=1,\cdot,J}^{l=1,\cdots,L}$ are obtained by multilevel aggregation applied to the spatial view $\mathbf{H}_\text{S}$ of the input contextual representations $\mathbf{H}$, and $\alpha_j^l$ are joint-specific influence scores obtained through cross-level update applied to $\{\mathbf{y}_j^l\}$, with compression matrix $\mathbf{W}^\text{compress}$ shared between spatial and temporal views.
Compared to simply averaging the outputs of multiple levels, the dynamic context-aware update offers more adaptability, leading to better performance, as shown in Table~\ref{tab:abla:cross level update}.

\begin{table*}[tp]
    \centering
    \caption{Results on both pose-based and mesh-based tasks across 3 in-domain datasets: AMASS~\cite{mahmood2019amass}, Human3.6M~\cite{ionescu2013h36m}, and FreeMan~\cite{wang2024freeman}.
    $\dagger$ indicates re-implementation to align with the setting of unified cross-domain 3D human motion modeling, i.e., a single training process on all tasks and datasets with a fully unified model.
    Lower numbers indicate better performance.
    }
    \vspace{-0.5em}
    \begin{subtable}[t]{0.99\linewidth}
        \centering
        \scriptsize
        \setlength\tabcolsep{0.4mm}
        \renewcommand\arraystretch{1.1}
        \vspace{-0.5em}
        \caption{\textbf{AMASS}~\cite{mahmood2019amass}}
        \vspace{-0.5em}
        \begin{tabular*}{\textwidth}{@{\extracolsep\fill}lccccccccccc@{}}
            \toprule
            \multirow{2}{*}{Models}     & \multirow{2}{*}{Venue} & \begin{tabular}[c]{@{}c@{}}Pose\\ Estimation \end{tabular} & \begin{tabular}[c]{@{}c@{}}Future Pose\\ Estimation \end{tabular} & \begin{tabular}[c]{@{}c@{}}Joint\\ Completion \end{tabular} & \begin{tabular}[c]{@{}c@{}}Motion\\ Prediction \end{tabular} & \begin{tabular}[c]{@{}c@{}}Motion\\ In-Between \end{tabular} & \begin{tabular}[c]{@{}c@{}}Mesh\\ Recovery \end{tabular} & \begin{tabular}[c]{@{}c@{}}Future Mesh\\ Recovery \end{tabular} & \begin{tabular}[c]{@{}c@{}}Joint\\ Completion \end{tabular} & \begin{tabular}[c]{@{}c@{}}Motion\\ Prediction \end{tabular} & \begin{tabular}[c]{@{}c@{}}Motion\\ In-Between \end{tabular} \\ \cmidrule{3-7}\cmidrule{8-12}
                &    & \multicolumn{5}{c}{Pose (MPJPE$\downarrow$)} & \multicolumn{5}{c}{Mesh (MPVE$\downarrow$)} \\ \midrule
            MotionBERT$^{\dagger}$~\cite{zhu2023motionbert}	&ICCV'23	&54.27 	&68.76 	&56.40 	&36.98 	&34.46 	&72.07 	&83.94 	&65.58 	&72.18 	&53.30 	\\
            PoseRetNet$^{\dagger}$~\cite{zheng2025poseretnet}	&ECCV'24	&40.58 	&46.74 	&64.42 	&39.47 	&33.58 	&478.51 	&481.95 	&79.86 	&67.14 	&59.85 	\\
            TCPFormer$^{\dagger}$~\cite{liu2025tcpformer}	&AAAI'25	&71.88 	&81.13 	&70.92 	&56.95 	&55.67 	&72.20 	&84.89 	&75.40 	&58.51 	&53.86 	\\
            HoT$^{\dagger}$~\cite{li2024hot}	&CVPR'24	&39.06 	&59.31 	&67.78 	&38.33 	&30.36 	&65.46 	&86.01 	&76.24 	&61.47 	&48.03 	\\
            PiC$^{\dagger}$~\cite{wang2024sic}	&CVPR'24	&32.65 	&42.68 	&50.94 	&32.19 	&25.39 	&43.34 	&56.35 	&63.59 	&48.57 	&38.64 	\\
            HiC	&Ours'25	&\textbf{24.96 }	&\textbf{34.06 }	&\textbf{41.93 }	&\textbf{24.98 }	&\textbf{17.23 }	&\textbf{38.03 }	&\textbf{48.98 }	&\textbf{55.90 }	&\textbf{44.42 }	&\textbf{34.44 }\\\toprule
            \end{tabular*}
    \end{subtable}
    
    \begin{subtable}[tp]{0.99\linewidth}
        \centering
        \scriptsize
        \setlength\tabcolsep{0.4mm}
        \renewcommand\arraystretch{1.1}
        \caption{\textbf{Human3.6M}~\cite{ionescu2013h36m}}
        \vspace{-0.5em}
        \begin{tabular*}{\textwidth}{@{\extracolsep\fill}lccccccccccc@{}}
            \toprule
            \multirow{2}{*}{Models}     & \multirow{2}{*}{Venue} & \begin{tabular}[c]{@{}c@{}}Pose\\ Estimation \end{tabular} & \begin{tabular}[c]{@{}c@{}}Future Pose\\ Estimation \end{tabular} & \begin{tabular}[c]{@{}c@{}}Joint\\ Completion \end{tabular} & \begin{tabular}[c]{@{}c@{}}Motion\\ Prediction \end{tabular} & \begin{tabular}[c]{@{}c@{}}Motion\\ In-Between \end{tabular} & \begin{tabular}[c]{@{}c@{}}Mesh\\ Recovery \end{tabular} & \begin{tabular}[c]{@{}c@{}}Future Mesh\\ Recovery \end{tabular} & \begin{tabular}[c]{@{}c@{}}Joint\\ Completion \end{tabular} & \begin{tabular}[c]{@{}c@{}}Motion\\ Prediction \end{tabular} & \begin{tabular}[c]{@{}c@{}}Motion\\ In-Between \end{tabular} \\ 
            \cmidrule{3-7}\cmidrule{8-12}
                &    & \multicolumn{5}{c}{Pose (MPJPE$\downarrow$)} & \multicolumn{5}{c}{Mesh (MPVE$\downarrow$)} \\ \midrule
            MotionBERT$^{\dagger}$~\cite{zhu2023motionbert}	&ICCV'23	&98.36 	&162.48 	&83.72 	&103.95 	&47.03 	&108.82 	&201.50 	&87.25 	&134.59 	&59.14 	\\
            PoseRetNet$^{\dagger}$~\cite{zheng2025poseretnet}	&ECCV'24	&96.33 	&135.22 	&93.93 	&103.41 	&54.56 	&328.24 	&343.60 	&104.39 	&136.80 	&67.01 	\\
            TCPFormer$^{\dagger}$~\cite{liu2025tcpformer}	&AAAI'25	&95.53 	&150.46 	&87.87 	&105.41 	&76.56 	&118.89 	&192.58 	&101.04 	&141.26 	&90.72 	\\
            HoT$^{\dagger}$~\cite{li2024hot}	&CVPR'24	&74.69 	&119.40 	&78.48 	&92.05 	&41.42 	&112.45 	&188.93 	&92.42 	&130.80 	&58.16 	\\
            PiC$^{\dagger}$~\cite{wang2024sic}	&CVPR'24	&74.35 	&116.86 	&76.10 	&91.14 	&37.83 	&95.38 	&155.42 	&82.02 	&127.54 	&50.71 	\\
            HiC	&Ours'25	&\textbf{62.94 }	&\textbf{115.79 }	&\textbf{58.37 }	&\textbf{87.81 }	&\textbf{24.93 }	&\textbf{78.67 }	&\textbf{151.53 }	&\textbf{69.52 }	&\textbf{125.67 }	&\textbf{42.89 }
	\\\toprule
            \end{tabular*}
    \end{subtable}
    \vspace{0.5em}

    \begin{subtable}[p]{0.99\linewidth}
        \centering
        \scriptsize
        \setlength\tabcolsep{1.5mm}
        \renewcommand\arraystretch{1.1}
        \vspace{-0.5em}
        \caption{\textbf{FreeMan}~\cite{wang2024freeman}}
        \vspace{-0.5em}
        \begin{tabular*}{\textwidth}{@{\extracolsep\fill}lccccccc@{}}
            \toprule
            \multirow{2}{*}{Models}     & \multirow{2}{*}{Venue}  & \begin{tabular}[c]{@{}c@{}}Joint\\ Completion \end{tabular} & \begin{tabular}[c]{@{}c@{}}Motion\\ Prediction \end{tabular} & \begin{tabular}[c]{@{}c@{}}Motion\\ In-Between \end{tabular}   & \begin{tabular}[c]{@{}c@{}}Joint\\ Completion \end{tabular} & \begin{tabular}[c]{@{}c@{}}Motion\\ Prediction \end{tabular} & \begin{tabular}[c]{@{}c@{}}Motion\\ In-Between \end{tabular} \\ \cmidrule{3-5} \cmidrule{6-8}
                &    & \multicolumn{3}{c}{Pose (MPJPE$\downarrow$)} & \multicolumn{3}{c}{Mesh (MPVE$\downarrow$)} \\ \midrule
            MotionBERT$^{\dagger}$~\cite{zhu2023motionbert}	&ICCV'23	 &113.29 	&99.22 	&61.33 	&112.97 	&130.24 	&68.74 	\\
            PoseRetNet$^{\dagger}$~\cite{zheng2025poseretnet}	&ECCV'24 	&141.43 	&120.19 	&88.40  	&137.64 	&142.09 	&89.12 	\\
            TCPFormer$^{\dagger}$~\cite{liu2025tcpformer}	&AAAI'25 	&134.54 	&128.35 	&107.46  	&186.41 	&201.44 	&168.54 	\\
            HoT$^{\dagger}$~\cite{li2024hot}	&CVPR'24 	&124.20 	&103.31 	&70.23 	&127.39 	&133.17 	&72.19 	\\
            PiC$^{\dagger}$~\cite{wang2024sic}	&CVPR'24 	&91.57 	&90.25 	&48.69 	&85.30 	&117.73 	&41.78 	\\
            HiC	&Ours'25	&\textbf{82.01 }	&\textbf{82.21 }	&\textbf{37.35 }	&\textbf{83.33 }	&\textbf{112.89 }	&\textbf{38.86 }
	\\\toprule
            \end{tabular*}
    \end{subtable}
\label{tab:in_domain_results}
\end{table*}

\begin{table*}[tp]
    \centering
    \scriptsize
    \setlength\tabcolsep{0.4mm}
    \renewcommand\arraystretch{1.1}
    \vspace{-1em}
    \caption{
    Results on 3DPW~\cite{von2018_3dpw} dataset. To evaluate out-of-domain generalization ability, all models are trained on 3 in-domain datasets (Human3.6M, AMASS, and FreeMan), and then tested on 3DPW dataset. 
    Lower numbers indicate better performance.
    }
    \vspace{-1em}
    \begin{tabular*}{\textwidth}{@{\extracolsep\fill}lccccccccccc@{}}
    \toprule
        \multirow{2}{*}{Models}     & \multirow{2}{*}{Venue} & \makecell{Pose\\ Estimation} & \makecell{Future Pose\\ Estimation} & \makecell{Joint\\ Completion} & \makecell{Motion\\ Prediction} & \makecell{Motion\\ In-Between} & \makecell{Mesh\\ Recovery} & \makecell{Future Mesh\\ Recovery} & \makecell{Joint\\ Completion} & \makecell{Motion\\ Prediction} & \makecell{Motion\\ In-Between} \\  \cmidrule{3-7}\cmidrule{8-12}
        &    & \multicolumn{5}{c}{Pose (MPJPE$\downarrow$)} & \multicolumn{5}{c}{Mesh (MPVE$\downarrow$)} \\ \midrule
        MotionBERT$^{\dagger}$~\cite{zhu2023motionbert}	&ICCV'23	&118.83 	&140.74 	&77.83 	&71.64 	&44.48 	&145.42 	&171.86 	&94.95 	&97.28 	&56.10 	\\
        PoseRetNet$^{\dagger}$~\cite{zheng2025poseretnet}	&ECCV'24	&120.88 	&132.30 	&81.91 	&71.20 	&48.94 	&314.26 	&311.91 	&97.30 	&95.12 	&68.59 	\\
        TCPFormer$^{\dagger}$~\cite{liu2025tcpformer}	&AAAI'25	&126.25 	&139.09 	&90.38 	&84.52 	&71.20 	&136.54 	&155.79 	&100.98 	&97.72 	&71.68 	\\
        HoT$^{\dagger}$~\cite{li2024hot}	&CVPR'24	&107.95 	&126.62 	&73.31 	&66.57 	&40.73 	&134.14 	&159.96 	&96.14 	&97.84 	&62.51 	\\
        PiC$^{\dagger}$~\cite{wang2024sic}	&CVPR'24	&107.22 	&120.73 	&71.60 	&62.96 	&36.15 	&133.05 	&151.83 	&92.53 	&93.49 	&53.69 	\\
        HiC	&Ours'25	&\textbf{96.36 }	&\textbf{111.93 }	&\textbf{56.71 }	&\textbf{55.97 }	&\textbf{23.00 }	&\textbf{119.45 }	&\textbf{141.95 }	&\textbf{82.60 }	&\textbf{84.44 }	&\textbf{35.28 }
	\\\toprule
    \end{tabular*}
\label{tab:out_of_domain_results}
\vspace{-1em}
\end{table*}

\begin{figure*}[tp]
    \hsize=\textwidth
    \centering
    \includegraphics[width=0.99\textwidth]{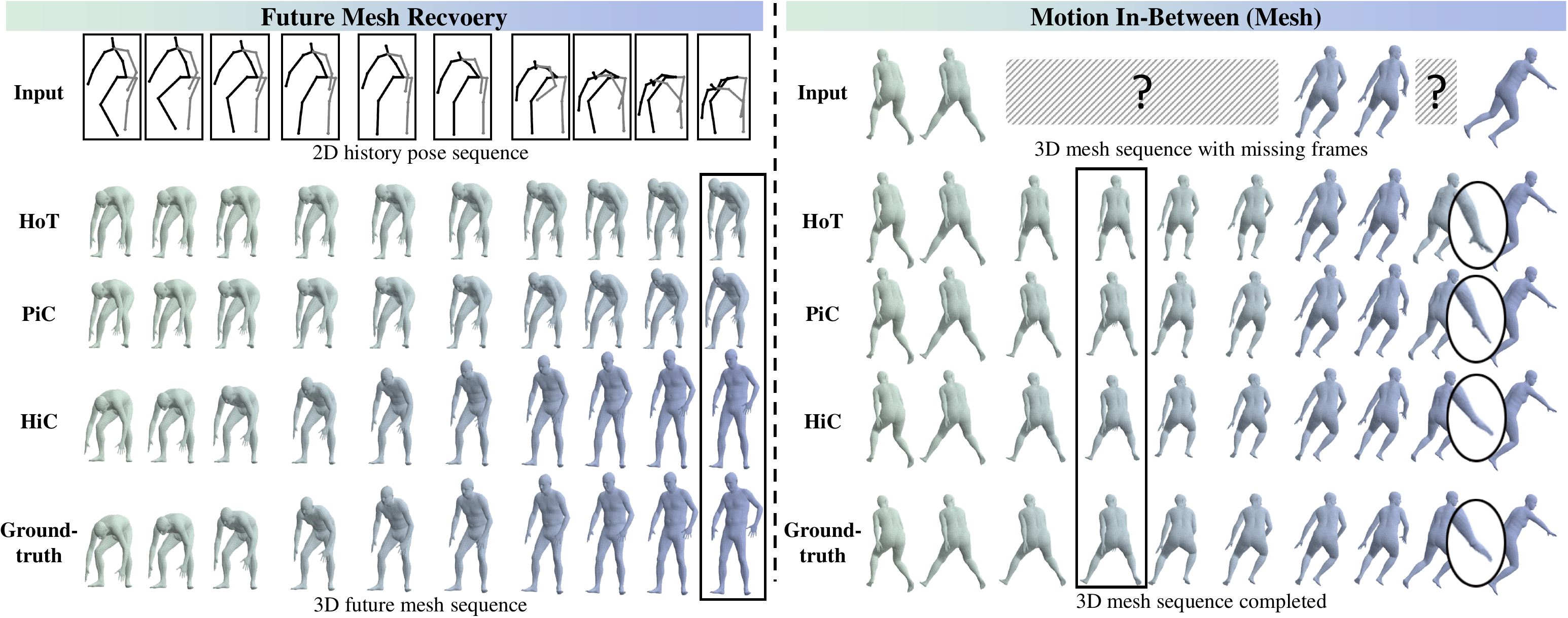}
    \caption{
    Qualitative results for future mesh recovery (left) and motion in-between (mesh) (right). Significant improvements are highlighted with boxes/circles, with details zoomed in for better clarity. Please refer to the supplementary material for additional qualitative results.
    }
    \label{fig:viz result}
\end{figure*}

\begin{table*}[t]
    \centering
    \setlength\tabcolsep{0.5mm}
    \renewcommand\arraystretch{1.1}
    \caption{Ablations on the number of anchors selected using max-min similarity prompt sampling. The models are trained on AMASS, Human3.6M, and FreeMan under the in-context setting, and then tested on AMASS and 3DPW.}
    \begin{tabular*}{\textwidth}{@{\extracolsep\fill}lccccccccccc@{}}
    \toprule[0.5mm]
    \multirow{2}{*}{Datasets} &\multirow{2}{*}{Anchors}  & \begin{tabular}[c]{@{}c@{}}Pose\\ Estimation \end{tabular} & \begin{tabular}[c]{@{}c@{}}Future Pose\\ Estimation \end{tabular} & \begin{tabular}[c]{@{}c@{}}Joint\\ Completion \end{tabular} & \begin{tabular}[c]{@{}c@{}}Motion\\ Prediction \end{tabular} & \begin{tabular}[c]{@{}c@{}}Motion\\ In-Between \end{tabular} & \begin{tabular}[c]{@{}c@{}}Mesh\\ Recovery \end{tabular} & \begin{tabular}[c]{@{}c@{}}Future Mesh\\ Recovery \end{tabular} & \begin{tabular}[c]{@{}c@{}}Joint\\ Completion \end{tabular} & \begin{tabular}[c]{@{}c@{}}Motion\\ Prediction \end{tabular} & \begin{tabular}[c]{@{}c@{}}Motion\\ In-Between \end{tabular} \\ \cmidrule{3-7}\cmidrule{8-12}
       & & \multicolumn{4}{c}{Pose (MPJPE$\downarrow$)} & \multicolumn{5}{c}{Mesh (MPVE$\downarrow$)} \\ \midrule
    \multirow{5}{*}{AMASS} &256 & 27.08 & 34.77 & 43.98 & 29.34 & 20.84 & 36.00 & 45.63 & 53.43 & 43.80 & 33.25 \\
     &512 & 27.20 & 34.71 & 42.71 & 27.83 & 19.63 & 37.42 & 46.81 & 54.67 & 43.92 & 33.22 \\
     &\cellcolor[gray]{0.8}800 &24.96 	&34.06 	&41.93 	&24.98 	&17.23 	&38.03 	&48.98 	&55.90 	&44.42 	&34.44 \\
     &1024 &26.01 	&35.43 	&44.40 	&26.38 	&18.87 	&39.95 	&51.00 	&57.41 	&45.10 	&35.43 
 \\ \midrule
    \multirow{4}{*}{3DPW} &256 & 104.99 & 119.82 & 62.54 & 63.85 & 27.40 & 126.37 & 149.06 & 87.57 & 108.29 & 37.35 \\
     &512 & 99.87 & 113.21 & 62.35 & 58.41 & 28.42 & 121.54 & 141.15 & 83.60 & 92.39 & 35.76 \\
     &\cellcolor[gray]{0.8}800 &96.36 	&111.93 	&56.71 	&55.97 	&23.00 	&119.45 	&141.95 	&82.60 	&84.44 	&35.28 \\
     &1024 &97.11 	&113.85 	&57.59 	&58.12 	&24.18 	&121.91 	&143.46 	&83.68 	&86.29 	&35.91 
 \\ \toprule[0.4mm]
    \end{tabular*}
\label{tab:abla:anchor num}
\vspace{-1em}
\end{table*}

\subsubsection{Context Injection}
In layer $k$, the query ($Q$) and prompt ($P$) branches each contain a \modelname~block to extract their respective contextual features, which can be written as:
\begin{equation}
        \mathbf{Z}_{Q}^{(k)}  = \textsc{XFusion}_{Q}^{(k)} (\mathbf{H_{Q}^{(k)}});
        \mathbf{Z}_{P}^{(k)}  = \textsc{XFusion}_{P}^{(k)} (\mathbf{H_{P}^{(k)}}),
\end{equation}
where $\mathbf{H_{Q}^{(k)}},\mathbf{H_{P}^{(k)}}\in\mathbb{R}^{F\times J\times H}$ are  query and prompt branches' respective inputs of the \modelname~block, and $ \mathbf{Z}_{Q}^{(k)}, \mathbf{Z}_{P}^{(k)}\in\mathbb{R}^{F\times J\times H'}$ are the outputs.
While the query and prompt branches gain deeper contextual understanding as their respective layers grow deeper, the query branch should be made aware of the contextual information from the prompts, as the query branch is the primary one to produce the final query output.
To this end, we apply a context injection module $\textsc{ContextInject}(\cdot)$ in each layer to inject a prompt context into the query branch, which drives it to learn the dependencies among the prompt and query at different depths. With the context injection module applied, the query and prompt branches' respective outputs in the current layer are obtained as:
\begin{equation}
        \mathbf{H}_{Q}^{(k+1)} = \textsc{ContextInject}(\mathbf{Z}_{P}^{(k)}, \mathbf{Z}_{Q}^{(k)});
         \mathbf{H}_{P}^{(k+1)} = \mathbf{Z}_{P}^{(k)},
\end{equation}
where $\mathbf{H}_{Q}^{(k+1)},\mathbf{H}_{P}^{(k+1)}$ indicate inputs of the next layer if there is one.
In HiC, the context injection module is implemented simply using a sum function as $ \mathbf{H}_{Q}^{(k+1)} = \mathbf{Z}_{P}^{(k)}+ \mathbf{Z}_{Q}^{(k)}$.

\subsubsection{Optimization}
To avoid any biased preference towards certain aggregation levels before optimization, the compression matrix $\mathbf{W}^\text{compress}$ and the bias term $[b^1,b^2,\cdots,b^L]$ are initialized with the following conditions:
\begin{equation}
    \begin{cases}
        \mathbf{W}^\text{compress} = \mathbf{0}; \\
        b^1 = b^2 = \cdots = b^L.
    \end{cases}
\end{equation}
These conditions ensure that before any optimization, the influence scores have the same values $\alpha_t^l=\frac{1}{L}$ for any frame $t$ and any level $l$, indicating an unbiased assumption of contributions of different aggregation levels prior to optimization.

During training, the loss functions involve computing pose joint positions and velocities, and also computing mesh vertex positions, joint rotation parameters, and shape parameters, following~\cite{zhu2023motionbert}. Please refer to the supplementary material for more network and optimization details.

\section{Experiments}
\label{sec:experiments}

\begin{figure*}[tp]
    \hsize=\textwidth
    \centering
    \includegraphics[width=0.99\textwidth]{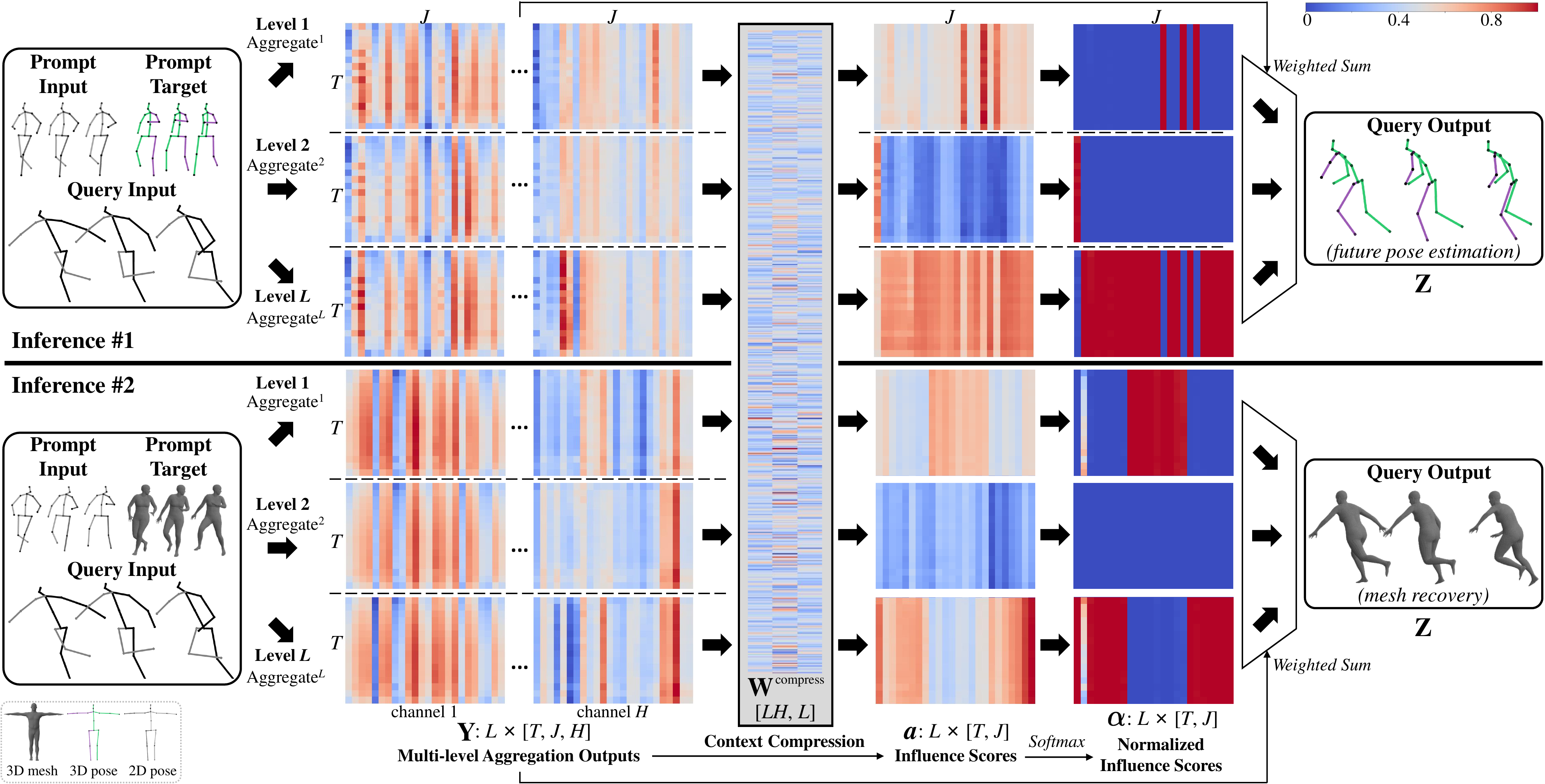}
    \vspace{-0.5em}
    \caption{
    Visualization of features and weights produced in two inference processes. In each inference, a different prompt is paired with the same query, and we visualize the resulting outputs from Multi-Level Aggregation, the compression weights, and the influence scores in Cross-Level Update. 
    It shows that, even for the same query, the multi-level aggregation and cross-level update in \model~can adapt to different prompts, learn dependencies through various levels of aggregation, spot the most influential contextual features both frame-wise as well as joint-wise, and generate the desired output dynamically under different contexts.
    It also verifies the necessity of aggregating contextual features at multiple levels and updating them by dynamically weighing the contributions across different levels.
    }
    \label{fig:viz mid feature}
\end{figure*}

\begin{figure*}[tp]
    \hsize=\textwidth
    \centering
    \includegraphics[width=0.99\textwidth]{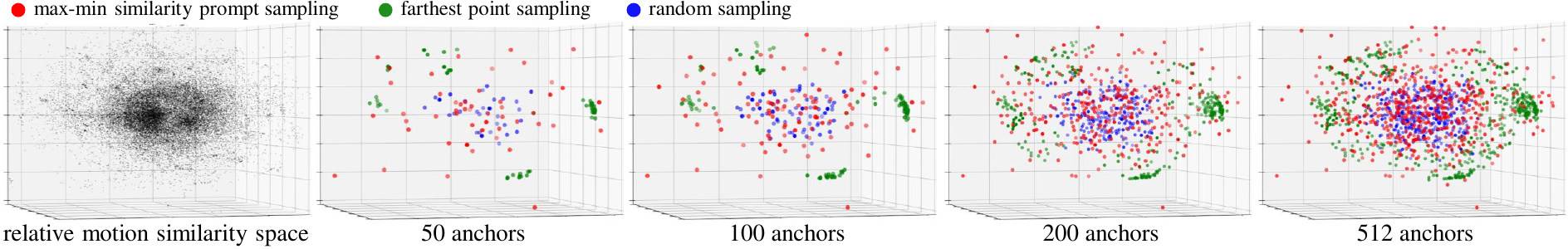}
    \vspace{-0.5em}
    \caption{
    Visual comparison between our proposed max-min similarity prompt sampling, farthest point sampling, and random sampling. Applied in the same relative motion similarity space, the FPS-based anchors (green) are concentrated in several peripheral regions, and the random-based anchors (blue) tend to gather in the densest regions (with overwhelmingly higher probabilities). In contrast, the SPS-based anchors (red) better capture the overall distributive pattern of all data, covering both peripheral and central, as well as dense and sparse regions, which is crucial for robust generalization across in-domain data and out-of-domain data.
    }
    \label{fig:viz sps fps}
\end{figure*}

\noindent\textbf{Implementation Details.}
We implement the proposed max-min similarity prompt sampling to obtain $\vert\mathcal{A}\vert=800$ anchors.
For the proposed model, we use $K=8$ layers and hidden feature dimension $H=128$. 
The motion sequences contain $F=16$ frames and $J=24$ joints. 
For tasks that require masks, we apply a 40\% mask ratio.
We implement Human-in-Context with PyTorch, with an AdamW optimizer with a linearly decaying learning rate, starting at 0.0002 and decreasing by 1\% after every epoch. 
We train Human-in-Context for 120 epochs on a Linux machine with 4 NVIDIA A6000 GPUs.

\noindent\textbf{Experimental Setting.}
    To validate the effectiveness of HiC as a unified cross-domain model, we compare it with a series of baseline methods, including MotionBERT~\cite{zhu2023motionbert}, PoseRetNet~\cite{zheng2025poseretnet}, TCPFormer~\cite{liu2025tcpformer}, HoT~\cite{li2024hot}, and PiC~\cite{wang2024sic}.
    All models are re-implemented to align with the setting of unified cross-domain 3D human motion modeling, i.e., one-time training for all tasks and datasets without any domain-specific model heads.

\noindent\textbf{Datasets, Tasks and Metrics.}
    For perfromance evalaution, we use 4 large-scale 3D human motion datasets, including AMASS~\cite{mahmood2019amass}, Human3.6M~\cite{ionescu2013h36m}, FreeMan~\cite{wang2024freeman}, and 3DPW~\cite{von2018_3dpw}.
    To validate both in-domain and out-of-domain generalization capability, we use AMASS, Human3.6M, and FreeMan as the in-domain datasets, while using 3DPW as the out-of-domain dataset.
    More specifically, we split each one of AMASS, Human3.6M, and FreeMan into a training set and a test set, in line with their respective splitting standards. 
    We use 3DPW as the test set.
    All models are trained on AMASS, Human3.6M, and FreeMan, and then tested on all four datasets.
    Each model is trained and tested on up to 10 tasks as described in 
    Table~\ref{tab:tasks}. 
    Among the 10 tasks, tasks with pose outputs are evaluated using Mean Per Joint Position Error (MPJPE), and tasks with mesh outputs are evaluated using Mean Per Vertex Error (MPVE), both in line with standard protocols.

\subsection{Quantitative Results}

\noindent\textbf{In-domain datasets.}
    Table~\ref{tab:in_domain_results} shows that the in-domain generalization capability is extensively evaluated using three datasets: AMASS~\cite{mahmood2019amass}, Human3.6M~\cite{ionescu2013h36m}, and FreeMan~\cite{wang2024freeman}. 
    After only a single training process on all three datasets, Human-in-Context outperforms all other methods consistently across all tasks among all datasets, whether pose- or mesh-based.

\begin{table*}[tp]
    \centering
    \setlength\tabcolsep{0.5mm}
    \renewcommand\arraystretch{1.1}
    \caption{Ablations on the soft anchor. The models are trained on AMASS, Human3.6M, and FreeMan under the in-context setting, and then tested on AMASS and 3DPW.}
    \vspace{-0.8em}
    \begin{tabular*}{\textwidth}{@{\extracolsep\fill}lccccccccccc@{}}
    \toprule[0.5mm]
    \multirow{2}{*}{Datasets} &\multirow{2}{*}{\#}  & \begin{tabular}[c]{@{}c@{}}Pose\\ Estimation \end{tabular} & \begin{tabular}[c]{@{}c@{}}Future Pose\\ Estimation \end{tabular} & \begin{tabular}[c]{@{}c@{}}Joint\\ Completion \end{tabular} & \begin{tabular}[c]{@{}c@{}}Motion\\ Prediction \end{tabular} & \begin{tabular}[c]{@{}c@{}}Motion\\ In-Between \end{tabular} & \begin{tabular}[c]{@{}c@{}}Mesh\\ Recovery \end{tabular} & \begin{tabular}[c]{@{}c@{}}Future Mesh\\ Recovery \end{tabular} & \begin{tabular}[c]{@{}c@{}}Joint\\ Completion \end{tabular} & \begin{tabular}[c]{@{}c@{}}Motion\\ Prediction \end{tabular} & \begin{tabular}[c]{@{}c@{}}Motion\\ In-Between \end{tabular} \\ \cmidrule{3-7}\cmidrule{8-12}
        &    & \multicolumn{5}{c}{Pose (MPJPE$\downarrow$)} & \multicolumn{5}{c}{Mesh (MPVE$\downarrow$)} \\ \midrule
    \multirow{2}{*}{AMASS} &w/o soft anchors &29.06 	&36.34 	&44.75 	&29.96 	&21.19 	&39.01 	&49.75 	&56.37 	&45.70 	&34.78  \\
     &\cellcolor[gray]{0.8}w soft anchors &24.96 	&34.06 	&41.93 	&24.98 	&17.23 	&38.03 	&48.98 	&55.90 	&44.42 	&34.44 
  \\ \midrule
    \multirow{2}{*}{3DPW} &w/o soft anchors &101.45 	&115.25 	&64.06 	&60.33 	&31.63 	&123.21 	&143.48 	&85.16 	&94.02 	&37.34 \\
     & \cellcolor[gray]{0.8}w/ soft anchors &96.36 	&111.93 	&56.71 	&55.97 	&23.00 	&119.45 	&141.95 	&82.60 	&84.44 	&35.28 
 \\
    \toprule[0.4mm]
    \end{tabular*}
\label{tab:abla:tup}
\vspace{-1em}
\end{table*}

\begin{table*}[tp]
    \centering
    \setlength\tabcolsep{0.5mm}
    \renewcommand\arraystretch{1.1}
    \caption{Ablations on the sampling method.
    The models are trained on AMASS, Human3.6M, and FreeMan under the in-context setting, and then tested on AMASS and 3DPW.}
    \vspace{-0.8em}
    \begin{tabular*}{\textwidth}{@{\extracolsep\fill}lccccccccccc@{}}
    \toprule[0.5mm]
    \multirow{2}{*}{Datasets} & \multirow{2}{*}{\#}  & \begin{tabular}[c]{@{}c@{}}Pose\\ Estimation \end{tabular} & \begin{tabular}[c]{@{}c@{}}Future Pose\\ Estimation \end{tabular} & \begin{tabular}[c]{@{}c@{}}Joint\\ Completion \end{tabular} & \begin{tabular}[c]{@{}c@{}}Motion\\ Prediction \end{tabular} & \begin{tabular}[c]{@{}c@{}}Motion\\ In-Between \end{tabular} & \begin{tabular}[c]{@{}c@{}}Mesh\\ Recovery \end{tabular} & \begin{tabular}[c]{@{}c@{}}Future Mesh\\ Recovery \end{tabular} & \begin{tabular}[c]{@{}c@{}}Joint\\ Completion \end{tabular} & \begin{tabular}[c]{@{}c@{}}Motion\\ Prediction \end{tabular} & \begin{tabular}[c]{@{}c@{}}Motion\\ In-Between \end{tabular} \\ \cmidrule{3-7}\cmidrule{8-12}
        &   & \multicolumn{5}{c}{Pose (MPJPE$\downarrow$)} & \multicolumn{5}{c}{Mesh (MPVE$\downarrow$)} \\ \midrule
    \multirow{4}{*}{AMASS} & random &32.20 	&43.51 	&52.67 	&36.62 	&33.30 	&44.76 	&54.53 	&61.92 	&47.64 	&36.54  \\
     & FPS &30.86 	&40.33 	&49.83 	&33.97 	&24.67 	&42.71 	&52.32 	&60.41 	&46.78 	&35.79   \\
     & cluster &29.15 	&40.06 	&45.79 	&28.53 	&21.41 	&43.03 	&55.85 	&60.50 	&49.60 	&39.35 
   \\
     &  \cellcolor[gray]{0.8}SPS &24.96 	&34.06 	&41.93 	&24.98 	&17.23 	&38.03 	&48.98 	&55.90 	&44.42 	&34.44 
  \\ \midrule
    \multirow{4}{*}{3DPW} & random &105.09 	&118.80 	&70.99 	&67.20 	&38.64 	&132.57 	&158.67 	&91.51 	&92.39 	&48.24  \\
     & FPS &104.35 	&118.45 	&68.47 	&65.49 	&37.58 	&127.45 	&156.90 	&88.41 	&92.25 	&44.24   \\
     & cluster &99.63 	&115.61 	&63.01 	&59.66 	&26.09 	&123.72 	&148.19 	&87.09 	&88.12 	&41.66 
   \\
     & \cellcolor[gray]{0.8}SPS &96.36 	&111.93 	&56.71 	&55.97 	&23.00 	&119.45 	&141.95 	&82.60 	&84.44 	&35.28 
  \\
    \toprule[0.4mm]
    \end{tabular*}
    \vspace{-0.8em}
\label{tab:abla:sampling method}
\end{table*}

    \textbf{1)}  Pose-based tasks include pose estimation, future pose estimation, joint completion (pose), motion prediction (pose), and motion in-between (pose). 
    The evaluation metric for these tasks is Mean Per Joint Position Error in millimeters, which measures the average distance between the output joint positions and the ground truth after aligning the root joint.
    For Human3.6M, HiC achieves state-of-the-art across all tasks. For pose estimation, it outperforms MotionBERT, PoseRetNet, TCPFormer, and PiC with a score of 62.94 mm. Similarly, for future pose estimation, HiC scores 115.79 mm. 
    For joint completion, motion prediction, and motion in-between for pose-based tasks, HiC also consistently achieves better results than other models. For motion in-between (pose), the MPJPE of HiC is 24.93 mm, much lower than MotionBERT at 47.03 mm.
    Similarly, HiC performs well for all tasks on AMASS. 
    For pose estimation, HiC achieves lower MPJPE (24.96 mm) than other methods, e.g., MotionBERT (54.27 mm) and PoseRetNet (40.58 mm).
    In addition, it performs well for future pose estimation (34.06 mm), joint completion (41.93 mm), and motion prediction (24.98 mm). HiC performs the best in motion in-between (pose), achieving 17.23 mm.
    For FreeMan, HiC performs favorably against other approaches for all applicable tasks. 

    \textbf{2)}  Mesh-based tasks include mesh recovery, future mesh recovery, joint completion (mesh), motion prediction (mesh), and motion in-between (mesh). The evaluation metric for these tasks is Mean Per Vertex Error in millimeters, which measures the average distance between the estimated and ground-truth vertices after aligning the root joint.
    In Human3.6M, HiC achieves 78.67 mm for mesh recovery, which is significantly better than other methods. 
    For future mesh recovery, HiC achieves 151.53 mm, outperforming all other models. 
    For joint completion and motion prediction (mesh), HiC shows lower errors compared to other methods.
    For AMASS, HiC achieves 38.03 mm for mesh recovery and 48.98 mm for future mesh recovery, outperforming all other models. 
    HiC also performs well in motion prediction (mesh) and joint completion (mesh). 
    For FreeMan, HiC also achieves the best results in all applicable tasks.

\noindent\textbf{Out-of-domain datasets.}
    As Table~\ref{tab:out_of_domain_results} shows, the out-of-domain generalization capability is evaluated on the 3DPW~\cite{von2018_3dpw} dataset. After a single training process on three in-domain datasets, Human-in-Context can effectively generalize to out-of-domain data, outperforming all other methods across all pose-based and mesh-based tasks.
    On the 3DPW dataset, HiC outperformed all other methods in pose-based tasks. For pose estimation, HiC achieves 96.36 mm.
    For future pose estimation, HiC achieves 111.93 mm, which is lower than other models. 
    For joint completion (pose), HiC achieves 56.71 mm.
    For motion prediction (pose), HiC achieves 55.97 mm, which is better than other models. 
    For motion in-between (pose), HiC also performs favorably against other models. 
    For mesh recovery, HiC achieves 119.45 mm. 
    In future mesh recovery and joint completion (mesh), HiC also performs favorably against other approaches. 
    For motion prediction (mesh) and motion in-between (mesh), HiC achieves 84.44 mm and 35.28 mm, respectively.

\subsection{Qualitative Results}

\subsubsection{Mesh Recovery and Motion In-Between}
Figure~\ref{fig:viz result} presents qualitative results for two different tasks, Future Mesh Recovery on the left and Motion In-Between (Mesh) on the right. These tasks are evaluated using three different methods: HoT, PiC, and HiC, along with the ground truth for comparison.

\textbf{1)}  For the Future Mesh Recovery task, the input is historical 2D poses, and the output is future 3D meshes. The input is represented as stick figures in the first row. In the second row, the results from HoT are shown, where the mesh recovery exhibits some inaccuracies, particularly in the torso and limb areas, leading to an imperfect recovery compared to the Ground Truth. The third row shows the results by PiC.
Finally, the fourth row shows the results from HiC, which demonstrate the most accurate mesh recovery, with the human mesh closely resembling the ground truth, particularly in terms of the positioning of the joints and the overall human figure.

\textbf{2)}  For the Motion In-Between task, the goal is to complete missing frames in a 3D mesh sequence. 
The second, third, and fourth rows show the results from HoT, PiC, and HiC, respectively, for interpolating between the poses. 
HoT exhibits noticeable discrepancies in the motion of the limbs and the positioning of the joints. 
Similarly, PiC shows some inaccuracies in the motion transition, and certain parts of the body are not smoothly interpolated. 
In contrast, the fourth row showing HiC's results shows more accurate transitions between the two poses, with the limbs moving more naturally and in line with the ground truth.

\begin{table*}[t]
    \centering
    \setlength\tabcolsep{0.5mm}
    \renewcommand\arraystretch{1.1}
    \caption{Ablations on the cross-level context update. Static means averaging the outputs of all levels, as opposed to dynamically weighing their contributions in~\model.
    The models are trained on AMASS, Human3.6M, and FreeMan under the in-context setting, and then tested on AMASS and 3DPW.}
    \vspace{-0.8em}
    \begin{tabular*}{\textwidth}{@{\extracolsep\fill}lccccccccccc@{}}
    \toprule[0.5mm]
    \multirow{2}{*}{Datasets} & \multirow{2}{*}{\#}  & \begin{tabular}[c]{@{}c@{}}Pose\\ Estimation \end{tabular} & \begin{tabular}[c]{@{}c@{}}Future Pose\\ Estimation \end{tabular} & \begin{tabular}[c]{@{}c@{}}Joint\\ Completion \end{tabular} & \begin{tabular}[c]{@{}c@{}}Motion\\ Prediction \end{tabular} & \begin{tabular}[c]{@{}c@{}}Motion\\ In-Between \end{tabular} & \begin{tabular}[c]{@{}c@{}}Mesh\\ Recovery \end{tabular} & \begin{tabular}[c]{@{}c@{}}Future Mesh\\ Recovery \end{tabular} & \begin{tabular}[c]{@{}c@{}}Joint\\ Completion \end{tabular} & \begin{tabular}[c]{@{}c@{}}Motion\\ Prediction \end{tabular} & \begin{tabular}[c]{@{}c@{}}Motion\\ In-Between \end{tabular} \\ \cmidrule{3-7}\cmidrule{8-12}
        &   & \multicolumn{5}{c}{Pose (MPJPE$\downarrow$)} & \multicolumn{5}{c}{Mesh (MPVE$\downarrow$)} \\ \midrule
    \multirow{2}{*}{AMASS} & static &35.40 	&46.87 	&63.95 	&37.04 	&28.95 	&38.92 	&49.74 	&56.23 	&44.55 	&34.42 \\
     & \cellcolor[gray]{0.8}dynamic &24.96 	&34.06 	&41.93 	&24.98 	&17.23 	&38.03 	&48.98 	&55.90 	&44.42 	&34.44 
  \\ \midrule
    \multirow{2}{*}{3DPW} & static &100.68 	&114.10 	&65.29 	&59.98 	&30.55 	&122.16 	&145.82 	&84.62 	&93.07 	&36.27 \\
     & \cellcolor[gray]{0.8}dynamic &96.36 	&111.93 	&56.71 	&55.97 	&23.00 	&119.45 	&141.95 	&82.60 	&84.44 	&35.28 
  \\
    \toprule[0.4mm]
    \end{tabular*}
    \vspace{-0.8em}
\label{tab:abla:cross level update}
\end{table*}
\begin{table*}[t]
    \centering
    \setlength\tabcolsep{0.5mm}
    \renewcommand\arraystretch{1.1}
    \caption{Ablations on the number of model layers. The models are trained on AMASS, Human3.6M, and FreeMan under the in-context setting, and then tested on AMASS and 3DPW.}
    \vspace{-0.8em}
    \begin{tabular*}{\textwidth}{@{\extracolsep\fill}lcccccccccccc@{}}
    \toprule[0.5mm]
    \multirow{2}{*}{Datasets} & \multirow{2}{*}{Layers}  & \multirow{2}{*}{Params} & \begin{tabular}[c]{@{}c@{}}Pose\\ Estimation \end{tabular} & \begin{tabular}[c]{@{}c@{}}Future Pose\\ Estimation \end{tabular} & \begin{tabular}[c]{@{}c@{}}Joint\\ Completion \end{tabular} & \begin{tabular}[c]{@{}c@{}}Motion\\ Prediction \end{tabular} & \begin{tabular}[c]{@{}c@{}}Motion\\ In-Between \end{tabular} & \begin{tabular}[c]{@{}c@{}}Mesh\\ Recovery \end{tabular} & \begin{tabular}[c]{@{}c@{}}Future Mesh\\ Recovery \end{tabular} & \begin{tabular}[c]{@{}c@{}}Joint\\ Completion \end{tabular} & \begin{tabular}[c]{@{}c@{}}Motion\\ Prediction \end{tabular} & \begin{tabular}[c]{@{}c@{}}Motion\\ In-Between \end{tabular} \\ \cmidrule{4-8}\cmidrule{9-13}
        &  & & \multicolumn{5}{c}{Pose (MPJPE$\downarrow$)} & \multicolumn{5}{c}{Mesh (MPVE$\downarrow$)} \\ \midrule
    \multirow{3}{*}{AMASS} & 4 & 36.14M & 28.48 & 35.71 & 44.51 & 29.37 & 21.09 & 36.81 & 46.05 & 54.39 & 43.54 & 32.81 \\
      & \cellcolor[gray]{0.8}8 & 46.67M &24.96 	&34.06 	&41.93 	&24.98 	&17.23 	&38.03 	&48.98 	&55.90 	&44.42 	&34.44 
  \\
      & 10 & 51.74M & 26.15 & 33.40 & 42.22 & 26.66 & 18.34 & 36.91 & 45.46 & 55.04 & 44.47 & 33.58 \\ \midrule
    \multirow{3}{*}{3DPW} & 4 & 36.14M & 105.61 & 120.62 & 61.51 & 63.42 & 25.52 & 129.01 & 151.38 & 87.93 & 92.93 & 36.15 \\
     &\cellcolor[gray]{0.8}8 & 46.67M &96.36 	&111.93 	&56.71 	&55.97 	&23.00 	&119.45 	&141.95 	&82.60 	&84.44 	&35.28 
 \\
     &10 & 51.74M & 98.63 & 114.24 & 60.16 & 61.30 & 25.73 & 117.73 & 141.52 & 84.61 & 86.35 & 38.44 \\ \toprule[0.4mm]
    \end{tabular*}
\label{tab:abla:layer}
\vspace{-1em}
\end{table*}

\subsubsection{Visualization of Features and Weights}
Figure~\ref{fig:viz mid feature} shows features and weights learned during the multilevel aggregation process in \model. 
For each prompt-query pair, the figure shows the outputs corresponding to different levels of aggregation, the context compression process, and the influence scores at each level. 
The influence scores are normalized, and the output is generated by dynamically adapting the features from multiple levels of aggregation. The visualization demonstrates how the model can learn dependencies between different levels of aggregation and how the most contributing contextual features, both frame-wise and joint-wise, are selected and weighted during the cross-level update. 
These results highlight the ability of the model to generate accurate outputs for various tasks, such as future pose estimation and mesh recovery, under different contexts. The weighted sum of these features ultimately leads to the desired output, showcasing the flexibility of \model~in handling diverse input prompts and tasks.
In addition, these results demonstrate the necessity of aggregating contextual features at multiple levels and updating them by dynamically weighing the contributions across different levels.

\subsection{Ablation Study}

\subsubsection{Prompting Strategy}

\noindent\textbf{Sampling Method.}
    Table~\ref{tab:abla:sampling method} and Figure~\ref{fig:viz sps fps} present the quantitative and qualitative ablation studies on the sampling method.
    Aside from achieving better quantitative results, SPS more effectively captures the overall data distribution by visual comparison. 
    FPS is constrained to peripheral areas by sampling only the farthest point in each iteration. 
    Random sampling neglects data in sparse regions with significantly lower probability, which are necessary for out-of-domain generalization.
    Additionally, we also compare SPS with a cluster-based sampling method, where we use k-means to cluster motion sequences and use the cluster centroids as the anchors. As the table shows, SPS outperforms the cluster-based sampling method.
    A better reflection of the overall data distribution contributes to HiC's robust in-domain and out-of-domain generalization capabilities.

\noindent\textbf{Number of Anchors.}
    Table~\ref{tab:abla:anchor num} shows an ablation study on the number of anchors obtained by max-min similarity prompt sampling.
    In general, increasing the number of anchors leads to better performance, but as the number increases even further, the gains begin to diminish in several tasks.
    The optimal number (800) of anchors ensures both generalization effectiveness and computational efficiency without the risk of overfitting.

\noindent\textbf{Soft Anchors.}
    Table~\ref{tab:abla:tup} shows an ablation study on soft anchors by comparing models with and without soft anchors. Incorporating soft anchors consistently leads to better performance in terms of both in-domain and out-of-domain generalization.

\subsubsection{Network Architecture}

\begin{figure}[t]
    \hsize=\columnwidth
    \centering
    \includegraphics[width=0.99\columnwidth]{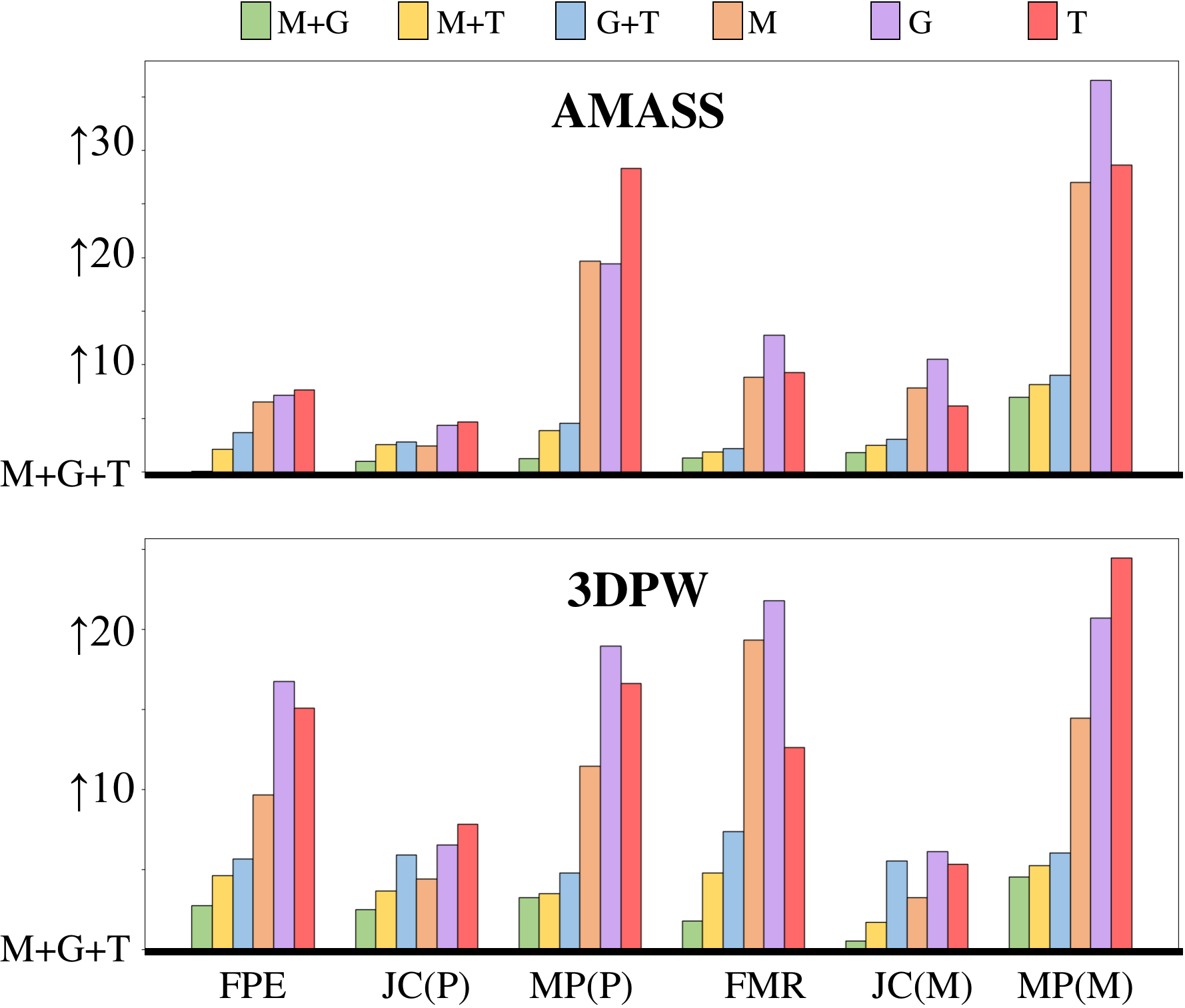}
    \vspace{-0.5em}
    \caption{
    Ablations on multi-level context update on AMASS (top) and 3DPW (bottom). Each bar corresponds to one of the various combinations of different levels, denoted by M (state-space model), T (self-attention), and G (graph convolution), respectively, for one of the 6 tasks.
    The results are reported relative to the default architecture (M+G+T). The longer the bar, the worse the performance relative to the default architecture (M+G+T).
    }
    \label{fig:abla:netarch}
    \vspace{-1em}
\end{figure}

\noindent\textbf{Multi-Level Context Aggregation.}   
    Figure~\ref{fig:abla:netarch} shows an ablation study on multi-level context aggregation by comparing different numbers of levels and different implementations of aggregation functions. We quantitatively analyze how state-space model (M), self-attention (T), and graph convolution (G) contribute to model performance.
    Each bar represents a specific module combination: M+G, M+T, G+T, M (state-space model only), G (graph convolution only), and T (self-attention only). Removing modules from the default architecture (M+G+T) consistently leads to a performance drop, highlighting the complementary nature of different aggregations.
    In addition to Figure~\ref{fig:viz mid feature}, Figure~\ref{fig:abla:netarch} further verifies the necessity of aggregating multi-level contextual features.

\noindent\textbf{Cross-Level Context Update.}   
    Table~\ref{tab:abla:cross level update} shows an ablation study on cross-level context update. Specifically, we compare the effectiveness of static and dynamic context updates at different levels in \model. The static context update method averages the outputs from all levels, while the dynamic context update method dynamically weighs the contributions of each level. The results demonstrate that the dynamic approach consistently outperforms the static one across all tasks and datasets.

\begin{figure}[t]
    \hsize=\columnwidth
    \centering
    \includegraphics[width=0.99\columnwidth]{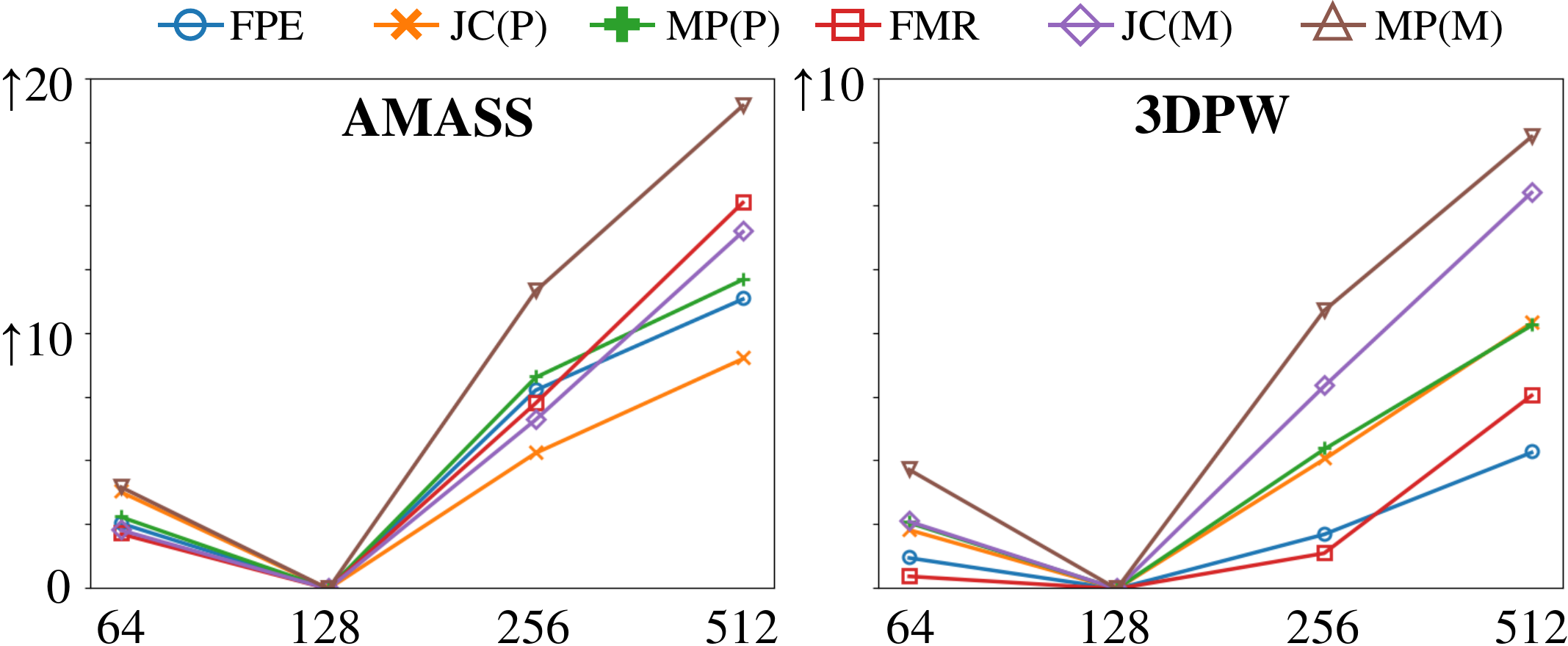}
    \caption{
    Ablations on feature dimension on AMASS (left) and 3DPW (right). Each line represents MPJPEs/MPVEs varying with different dimensions for one of the 6 tasks.
    The higher the points, the worse the performance relative to the default dimension (128).
    }
    \label{fig:abla:dim}
    \vspace{-1em}
\end{figure}

\noindent\textbf{Feature Dimension.}
    Figure~\ref{fig:abla:dim} shows an ablation study on feature dimensions.
    The model performs best with 128 feature dimensions, as they are sufficiently expressive without overcomplicating the representation. Larger dimensions lead to overparameterization, inefficiency, and reduced generalization. 
    Meanwhile, reducing the dimension also results in accuracy drops, as it is insufficient to capture the complex contextual dependencies.

\noindent\textbf{Number of Layers.}
    Table~\ref{tab:abla:layer} shows an ablation study on the number of model layers.
    Increasing the number of layers improves results across several tasks but also requires a bigger parameter size.
    After balancing effectiveness and efficiency, we use 8 layers to implement the proposed \model~in Human-in-Context.


\section{Future Work}

\textbf{Broader Domain Scope.} The exploration of unified 3D human motion modeling across domains could be expanded to a much broader scope of domains.
Regarding modalities, multi-modal data including RGB videos and point cloud sequences could be employed, providing a more comprehensive perspective on human appearance and depth details, which could benefit in-context learning by providing more sufficient and abundant contextual information.
Regarding tasks, future work could investigate incorporating additional complex tasks such as action recognition, person re-identification, and segmentation, which would further increase the model's versatility.
Furthermore, more datasets would improve generalization capabilities for increased data scale and motion diversity.
Datasets from different contextual backgrounds could further strengthen the robustness of the model, enabling its application to a wider array of real-world use cases such as robotics and augmented reality.

\noindent\textbf{Efficiency and Scalability.} 
In addition to expanding the scope, future work should also focus on improving the efficiency and scalability of models. Methods for reducing computational complexity and memory usage, such as using more efficient attention mechanisms or network pruning techniques, could help deploy cross-domain models in resource-constrained environments. 
Additionally, advancing transfer learning techniques could allow the model to better generalize from a smaller number of annotated data, speeding up the training process and making the model more accessible for practical applications. 

\noindent\textbf{Advanced Prompt Engineering.}
As models become more multi-modal (handling both 3D and 2D data), there would be a need for prompts that can efficiently fuse multiple data sources. Future work could generate prompts that integrate 3D, visual, and even textual cues to enable more robust and contextually aware models.
As models become more complex, understanding how a prompt influences the model's decision-making process will be critical. Future work could explore techniques to make prompt engineering more interpretable, helping developers understand which prompt aspects have the greatest impact on performance.

\noindent\textbf{Powerful Backbone.}
While the current backbone in Human-in-Context offers strong performance across various tasks, incorporating more advanced architectures could further enhance its ability to capture complex features and interactions within the data. Improving the backbone's capacity to handle more complicated features, interactions, and sparse data would not only enhance the overall performance but also provide better scalability for real-world applications across diverse scenarios, from robotics to augmented reality and beyond.

\section{Conclusion}
\label{sec:conclusion}
We present Human-in-Context (HiC) for cross-domain 3D human motion modeling. 
Unlike existing cross-domain models that rely on domain-specific components and multi-stage training, HiC can generalize across multiple domains, including cross-modality, cross-task, and cross-dataset scenarios, all with a unified model and a single training process.
HiC introduces a max-min similarity prompt sampling strategy to address the challenge of generalizing across diverse domains. 
Additionally, HiC consists of a \model~that captures in-context dependencies through multi-level aggregation and cross-level updates. 
Experiments across two modalities, ten tasks, and four datasets demonstrate that HiC outperforms our previous Pose-in-Context (PiC) by a large margin,  highlighting its effectiveness for both in- and out-of-domain generalization.

{
\normalem
\bibliographystyle{IEEEtran}
\bibliography{main}
\vspace{-4em}
}

\vspace{-5mm}
\begin{IEEEbiography}
[{\includegraphics[width=1in,height=1.25in,clip,keepaspectratio]{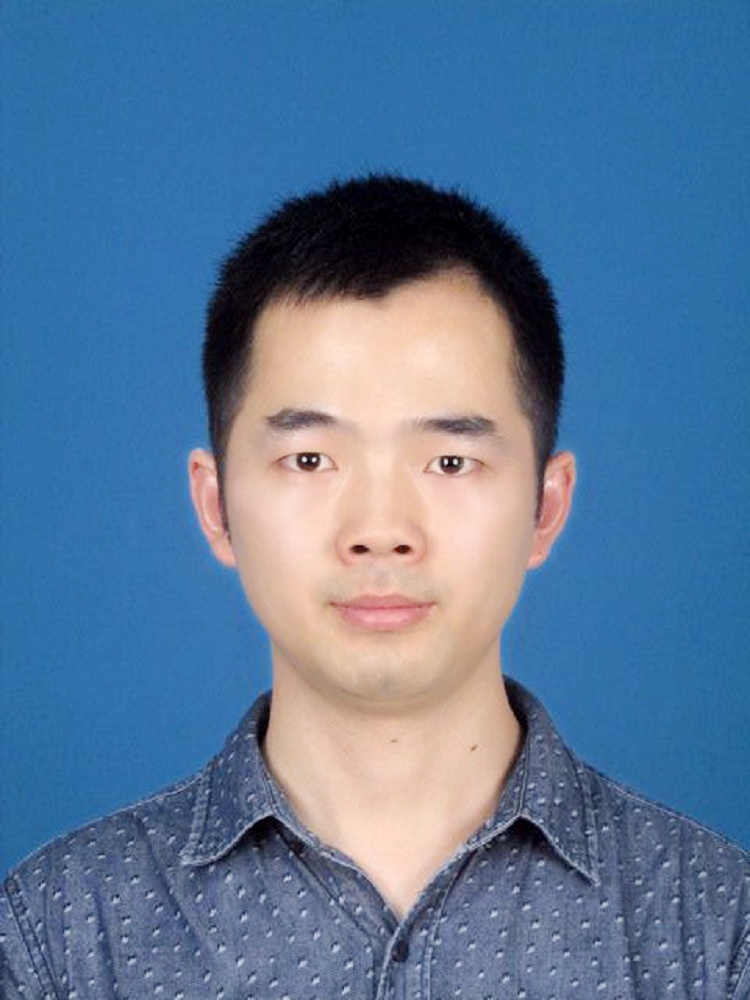}}]
{Mengyuan Liu} received his Ph.D. degree from the School of Electrical Engineering and Computer Science, Peking University, China. He was a research fellow at the School of Electrical and Electronic Engineering, Nanyang Technological University, Singapore. Currently, he is an Assistant Professor at Peking University, Shenzhen Graduate School. His research focuses primarily on human-centric perception and human-robot interaction. His work has been published in leading conferences and journals, including NeurIPS, CVPR, ICRA, T-IP, T-CSVT, T-MM, and PR. He actively serves as a reviewer for several top-tier international conferences and journals, such as NeurIPS, CVPR, T-IP, T-RO, and T-PAMI.
\end{IEEEbiography}

\vspace{-20mm}
\begin{IEEEbiography}
[{\includegraphics[width=1in,height=1.25in,clip,keepaspectratio]{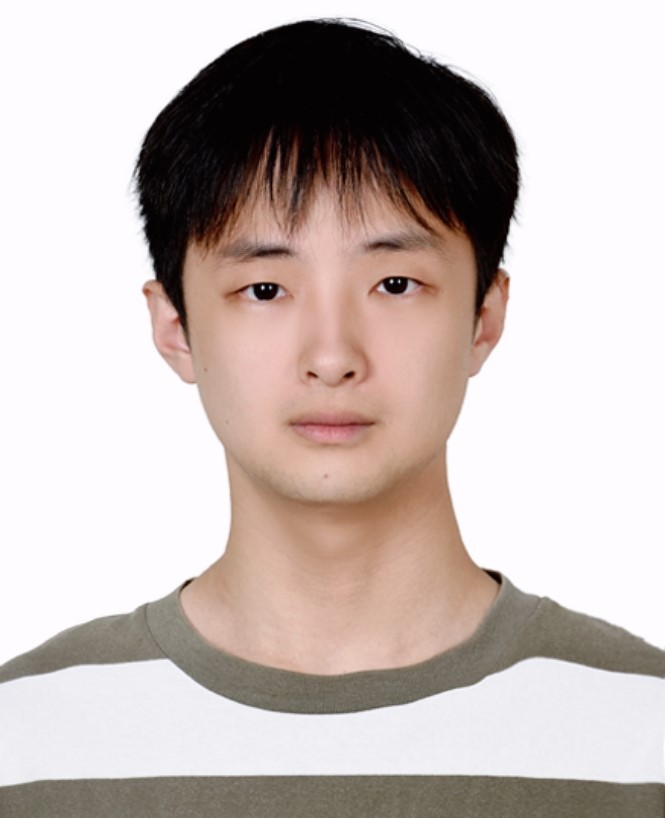}}]
{Xinshun Wang} is a Ph.D. student in computer science and technology at Peking University. He received the M.S. Degree from Sun Yat-sen University in 2024, and B.E. Degree from South China University of Technology in 2021. He has published papers in T-IP, CVPR, and AAAI. His research focuses on human-centric perception and generation. He is also interested in in-context learning and human foundation models. He has served as a reviewer for journals such as T-MM.
\end{IEEEbiography}

\vspace{-16mm}
\begin{IEEEbiography}
[{\includegraphics[width=1in,height=1.25in,clip,keepaspectratio]{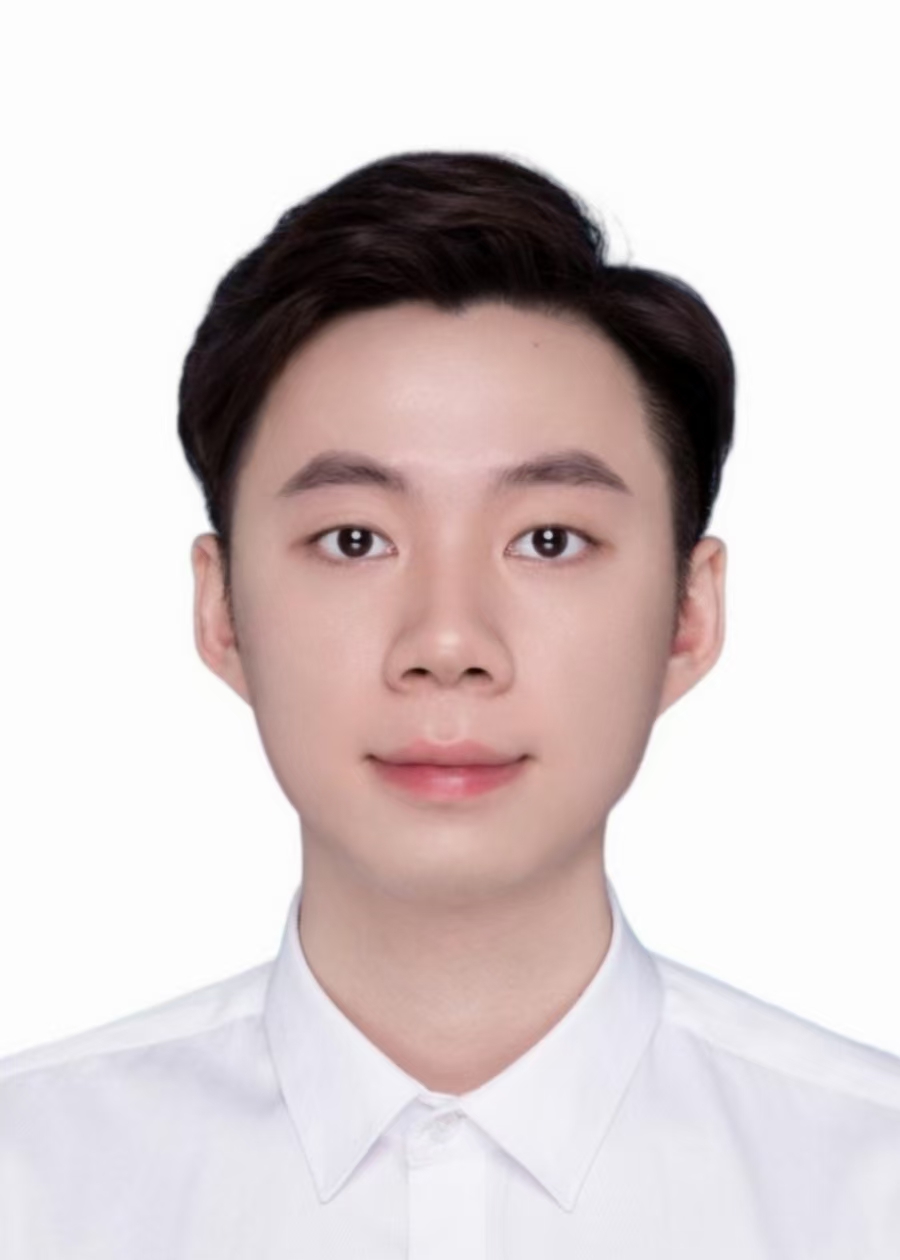}}]
{Zhongbin Fang} is a graduate student at Sun Yat-sen University and a research intern at Tencent. He has published papers in NeurIPS, CVPR, and CVIU. His research focuses on Deep Learning, 3D Point Cloud Analysis, In-Context Learning, and Video Generation. He is also highly interested in Large Language and Multimodal Models. He has served as a reviewer for journals including T-MM and T-CSVT.
\end{IEEEbiography}

\vspace{-16mm}
\begin{IEEEbiography}
[{\includegraphics[width=1in,height=1.25in,clip,keepaspectratio]{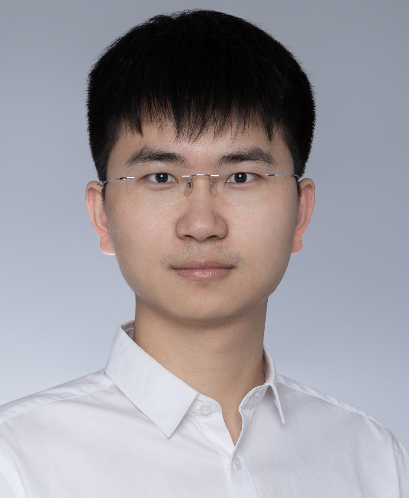}}]
{Deheng Ye} is the Director of AI Applications at Tencent, Shenzhen, China, where he spearheads a dynamic team of engineers and researchers in the development of large-scale machine learning platforms and intelligent AI agents. He also serves as the Deputy Director of the Ten\uline{cen}t-N\uline{TU} Joint \uline{R}esearch Laborator\uline{y} (CENTURY). His leadership has been instrumental in innovating and integrating pioneering AI technologies into real-world applications, with online AI service requests reaching billions per day worldwide. Deheng Ye completed his Ph.D. in Computer Science from Nanyang Technological University (NTU), Singapore, in 2016.
\end{IEEEbiography}

\vspace{-16mm}
\begin{IEEEbiography}
[{\includegraphics[width=1in,height=1.25in,clip,keepaspectratio]{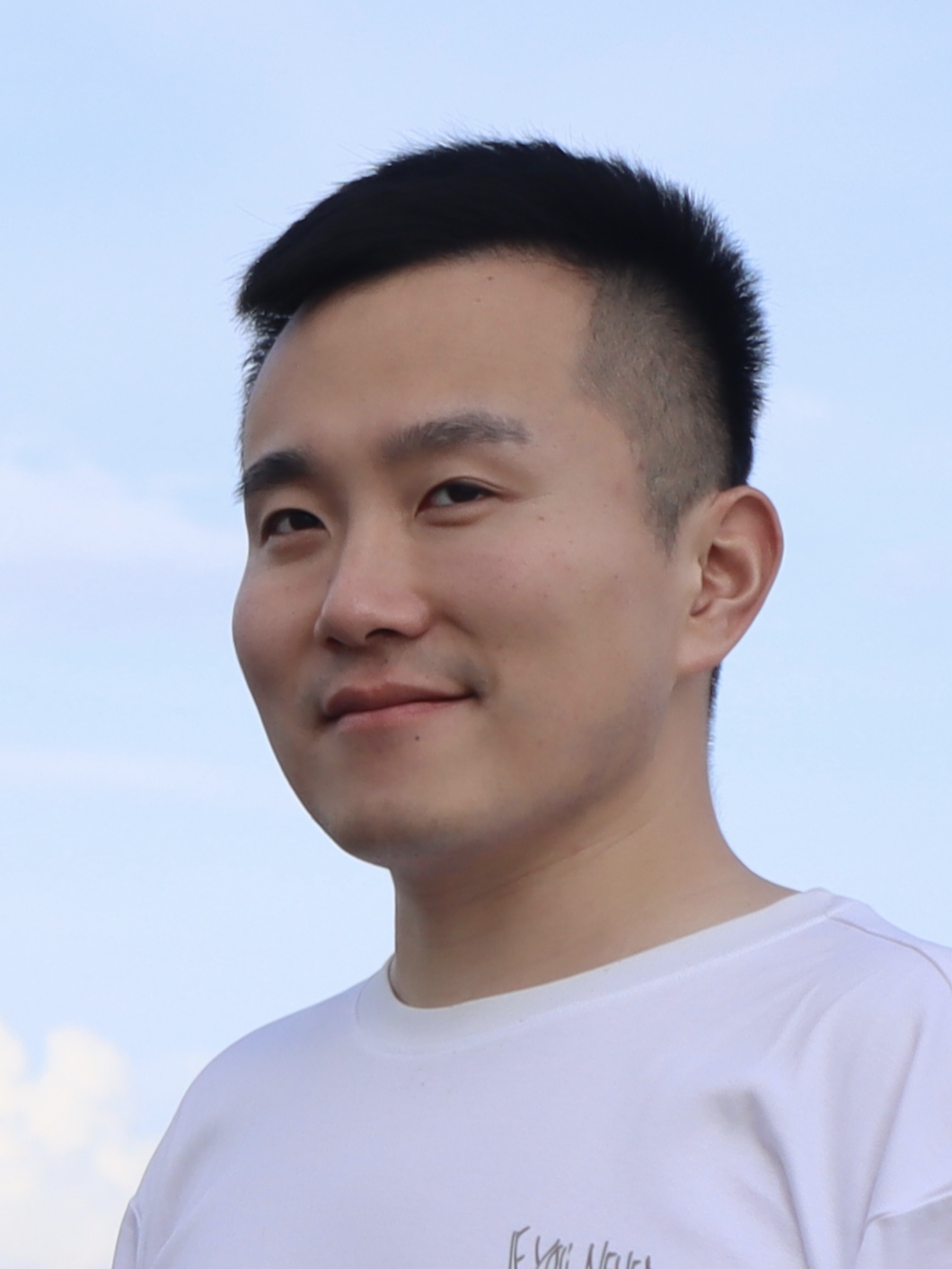}}]
{Xia Li} is a postdoc researcher at ETH Zurich. He received his master's degree in 2020 from Peking University and his doctoral degree from ETH Zurich. 
He is conducting research on the interdisciplinary area between image processing and radiotherapy, empowering personalized medicine with artificial intelligence. Mr. Li received the ICCR Rising Star award in 2024. He has published papers in top-tier conferences and journals such as CVPR, ICCV, ICLR, and NeurIPS.
\end{IEEEbiography}

\vspace{-16mm}
\begin{IEEEbiography}
[{\includegraphics[width=1in,height=1.25in,clip,keepaspectratio]{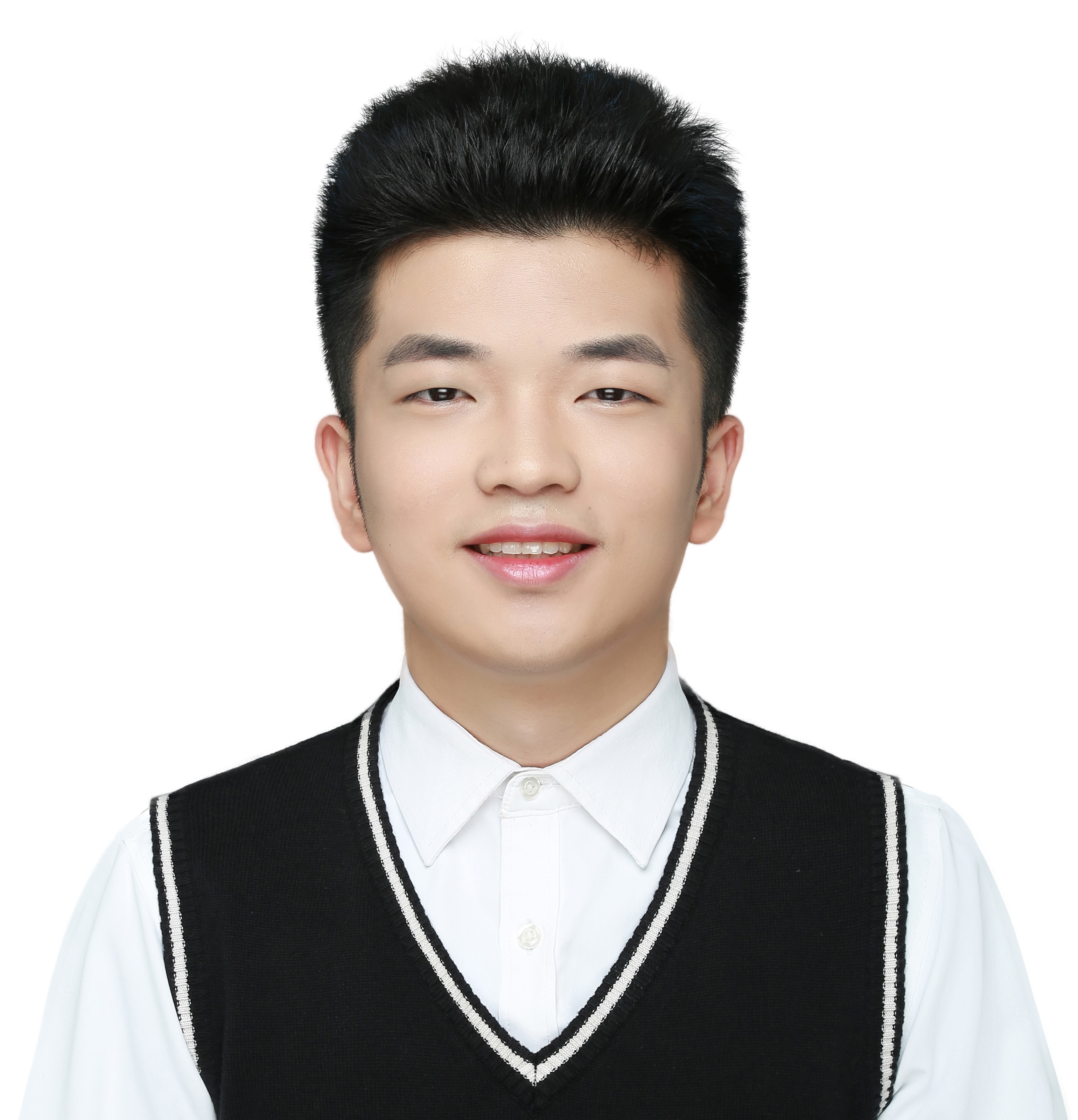}}]
{Tao Tang} received the BE degree from Central South University, Changsha, China, in 2023. He is currently working toward a postgraduate degree with the School of Electronic and Computer Engineering, Peking University, Shenzhen Graduate School, advised by Prof. Hong Liu. He has published papers in ACM MM and IROS. His current research interest lies in human mesh recovery.
\end{IEEEbiography}

\vspace{-16mm}
\begin{IEEEbiography}
[{\includegraphics[width=1in,height=1.25in,clip,keepaspectratio]{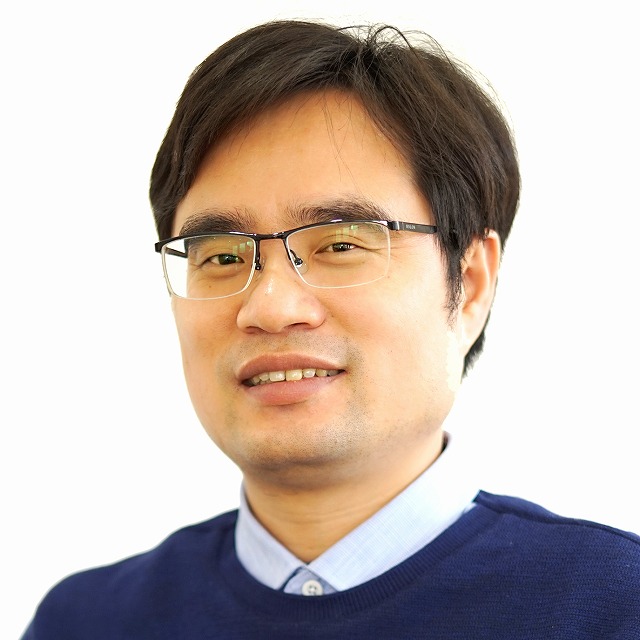}}]
{Songtao Wu} received the B.Sc. degree in electronic information engineering from Lanzhou University, Lanzhou, China, in 2009, the M.S. degree in circuits and systems from Peking University, Beijing, China, in 2012, and the Ph.D. degree from the Department of Computing, The Hong Kong Polytechnic University, Hong Kong, in 2017. He is currently a researcher with the AI Research Group, Beijing Lab, Sony R\&D Center. His research interests include human-computer interaction, generative AI, and multimedia security. 
\end{IEEEbiography}

\vspace{-16mm}
\begin{IEEEbiography}
[{\includegraphics[width=1in,height=1.25in,keepaspectratio]{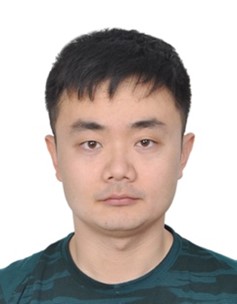}}]
{Xiangtai Li} is a Research Scientist at Bytedance, Singapore, working on multi-modal large language models. He was a Research Fellow at MMLab@NTU and a member of the Multimedia Laboratory at Nanyang Technological University. He received his Ph.D. degree from Peking University in 2022. His research interests include computer vision and machine learning with a focus on scene understanding, segmentation, video understanding, and multi-modal learning. He regularly reviews top-tier conferences and journals, including CVPR, ICCV, ICLR, ECCV, ICML, NeurIPS, T-PAMI, and IJCV. He serves as an area chair for ICCV, ICML, and ICLR. 
\end{IEEEbiography}

\begin{IEEEbiography}
[{\includegraphics[width=1in,height=1.25in,clip,keepaspectratio]{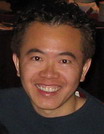}}] 
{Ming-Hsuan Yang} is a Professor of Electrical
Engineering and Computer Science at the University
of California, Merced. Yang serves as a program
co-chair of the ICCV in 2019. 
He received the Best Paper Award at ICML 2024, Longuet-Higgins Prize at CVPR 2023, Best Paper Honorable Mention at CVPR 2018, Nvidia Pioneer Research Award in 2018 and 2017, NSF CAREER award in 2012, and Google Faculty Award in 2009. He is a Fellow of the IEEE, ACM and AAAI.
\end{IEEEbiography}

\end{document}